%% file: main.tex
\documentclass[10pt,twocolumn,letterpaper]{article}

\usepackage{sample/iccv}
\usepackage{times}
\usepackage{epsfig}
\usepackage{graphicx}
\usepackage{amsmath}
\usepackage{amssymb}
\usepackage{subcaption}
\usepackage{booktabs,chemformula}
\usepackage{mathtools}
\usepackage{multirow}
\usepackage{balance}
\usepackage{algorithm}
\usepackage[algo2e,ruled,vlined]{algorithm2e}
\usepackage{listings}
\usepackage{color}
\usepackage{flafter}
\usepackage{float}
\usepackage{bbold}
\usepackage[pagebackref=true,breaklinks=true,letterpaper=true,colorlinks,bookmarks=false]{hyperref}

\iccvfinalcopy 

\def\iccvPaperID{6425} 

\ificcvfinal\fi

\begin{document}

\title{Elsa: Energy-based learning for semi-supervised anomaly detection}

\input{author}


\newcommand{\suw}[1]{\textcolor{red}{#1}}
\newcommand{\mc}[1]{\textcolor{magenta}{#1}}
\newcommand{\hyun}[1]{\textcolor{green!10!orange!90!}{#1}}
\newcommand{\sean}[1]{\textcolor{purple}{#1}}
\newcommand{\ryan}[1]{\textcolor{pink}{#1}}

\newcommand{\cutabstractup}{\vspace*{-0.2in}}
\newcommand{\cutabstractdown}{\vspace*{-0.2in}}
\newcommand{\cutsectionup}{\vspace*{-0.15in}}
\newcommand{\cutsectiondown}{\vspace*{-0.12in}}
\newcommand{\cutsubsectionup}{\vspace*{-0.15in}}
\newcommand{\cutsubsectiondown}{\vspace*{-0.07in}}
\newcommand{\cutparagraphup}{\vspace*{-0.15in}}

\newcommand{\model}{\textsf{Elsa}}

\maketitle
\ificcvfinal\thispagestyle{empty}\fi

\input{0_abstract}
\input{1_introduction}

\input{2_related_works}
\input{3_background}
\input{4_model}
\input{5_experiments}
\input{6_discussion}

\balance
{\small
\bibliographystyle{sample/ieee_fullname}
\bibliography{egbib}
}

\onecolumn
\newpage
\input{7_Appendix}

\end{document}

%% file: author.tex
\author{Sungwon Han\thanks{Equal contribution to this work.}\textsuperscript{\rm ~~1,2}, Hyeonho Song\footnotemark[1]\textsuperscript{\rm ~~1,2}, Seungeon Lee\textsuperscript{\rm 1,2}, Sungwon Park\textsuperscript{\rm 1,2}, Meeyoung Cha\textsuperscript{\rm 2,1} \\

\textsuperscript{\rm 1}School of Computing, KAIST, South Korea \\
\textsuperscript{\rm 2}Data Science Group, IBS, South Korea\\
\small{\{lion4151, hyun78, archon159, deu30303\}@kaist.ac.kr ~~\{mcha\}@ibs.re.kr}}

%% file: 0_abstract.tex
\begin{abstract}
Anomaly detection aims at identifying deviant instances from the normal data distribution. Many advances have been made in the field, including the innovative use of unsupervised contrastive learning. However, existing methods generally assume clean training data and are limited when the data contain unknown anomalies. This paper presents \model{}, a novel semi-supervised anomaly detection approach that unifies the concept of energy-based models with unsupervised contrastive learning. \model{} instills robustness against any data contamination by a carefully designed fine-tuning step based on the new energy function that forces the normal data to be divided into classes of prototypes. Experiments on multiple contamination scenarios show the proposed model achieves SOTA performance. Extensive analyses also verify the contribution of each component in the proposed model. Beyond the experiments, we also offer a theoretical interpretation of why contrastive learning alone cannot detect anomalies under data contamination.\looseness = -1
\end{abstract}

%% file: 1_introduction.tex
\section{Introduction}
\noindent
Anomaly detection~\cite{chandola2009survey}, also known as novelty detection~\cite{PIMENTEL2014215}, identifies out-of-distribution (OOD) instances from the predominant normal data. Conventional detection approaches model the probability distribution $p(\mathbf{x})$ of the normal data as \textit{normality score} implicitly or explicitly and identify deviant input with a small normality score. Various models have been used for estimating $p(\mathbf{x})$, including generative adversarial networks~\cite{schlegl2017unsupervised, zaheer2020old}, autoencoders~\cite{abati2019latent, choi2020novelty}, one-class classifiers~\cite{ruff2018deep,Ruff2020Deep}, and discriminative models with surrogate tasks~\cite{wang2019effective}.

\begin{figure}[t!]
\centering
\begin{subfigure}[t]{0.23\textwidth}
       \centering\includegraphics[width=\textwidth]{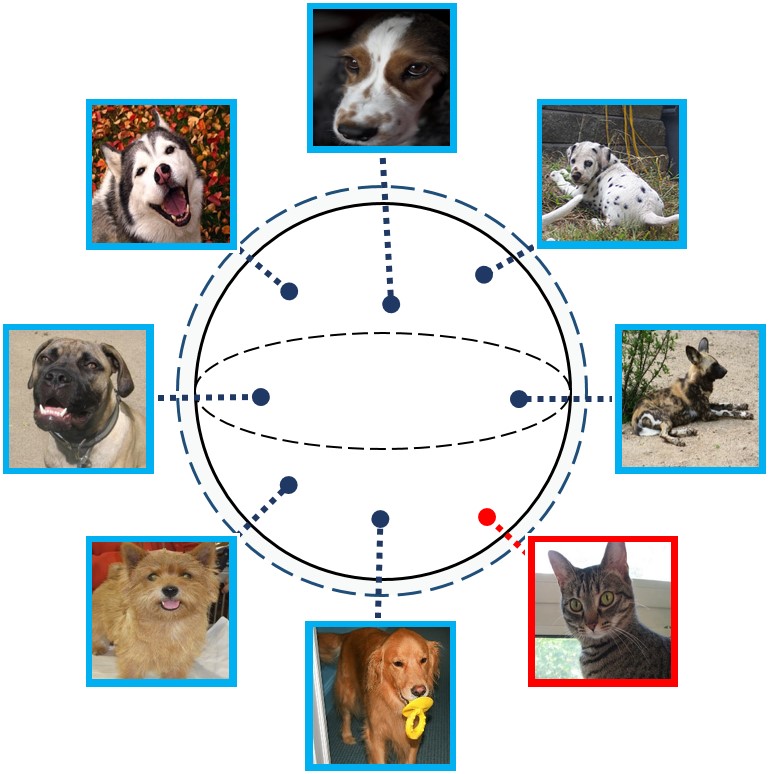}
      \caption{
      Embedding of CSI is a uniformly distributed hypersphere where anomalies (marked red) are hard to detect. \looseness=-1
      }
      \label{fig:intro_contrastive}
\end{subfigure}
\hspace{1mm}
\begin{subfigure}[t]{0.23\textwidth}
       \centering\includegraphics[width=\textwidth]{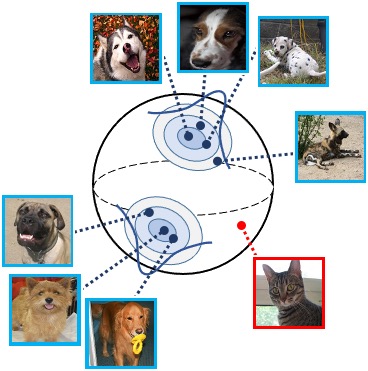} 
      \caption{
      Energy-based fine-tuning of \model{} embeds similar images nearby, leaving room for anomalies to be separated.
      }
      \label{fig:intro_ours}
\end{subfigure}
\caption{The motivating example comparing embeddings from existing unsupervised contrastive learning approaches and energy-based fine-tuning of the proposed model.}
\label{fig:motivate}
\end{figure}

Among the recent literature, CSI~\cite{tack2020csi} is the state-of-the-art novelty detection method based on contrastive learning. This model treats augmented input as positive samples and the distributionally-shifted input as negative samples, which leads to a substantial performance gain. Yet, it shares a common limitation with extant methods in that the model assumes clean training data and fails to learn $p(\mathbf{x})$ when the data contains unknown anomalies as in various real-world scenarios. This limitation occurs because CSI uses a hypersphere embedding space that is uniformly distributed~\cite{chen2020intriguing,wang2020understanding}. The uniformity makes it challenging to distinguish OOD samples. Fig.~\ref{fig:intro_contrastive} demonstrates this limitation. \looseness=-1

As a solution in this paper, we introduce \model{} (\textsf{E}nergy based \textsf{l}earning for \textsf{s}emi-supervised \textsf{a}nomaly detection), a novel anomaly detection method that unifies contrastive learning and energy-based functions. Our model benefits from the high representation power of unsupervised contrastive learning via its pre-training step, which can accommodate existing algorithms~\cite{chen2020simple,han2020mitigating,he2020momentum}. It then applies a carefully designed energy function over the pre-trained embedding to learn the probability distribution $p(\mathbf{x})$ of normal data, with the help of a small set of labels that indicate whether given samples are normal or OOD. The energy-based fine-tuning step embeds similar samples nearby and can distinguish OOD samples from the mostly normal data via the energy score: low-energy corresponds to compatible data distribution (e.g., dog images in Fig.~\ref{fig:intro_ours}), and high-energy represents incompatibility (e.g., a cat image in Fig.~\ref{fig:intro_ours}).  \looseness=-1

\model{}'s energy function does not require any explicit density estimator. Instead, it transforms an unsupervised contrastive problem into a non-parametric classification task by introducing the concept of \textit{prototypes}, where a prototype vector functions as a subclass that reflects the heterogeneity of the normal data distribution. The energy score is computed as logits from the discriminative classifier, which denote the cosine similarity between a data instance and prototypes. \model{} is structurally different from other energy-based models that require knowledge of the ground-truth class information~\cite{liu2020energy}. Furthermore, \model{}'s training is stable compared to other works that utilized energy functions as generative models~\cite{gao2020flow, grathwohl2019your}. \looseness=-1

\model{} shows nontrivial improvement over the best-performing models tested on CIFAR-10 and other benchmark datasets. Experiments confirm this gain is attributed to directly learning $p(\mathbf{x})$ via the energy function applied on unsupervised embedding learning. We empirically demonstrate the model's  robustness by considering three representative scenarios. The main highlights of this paper are as follows:
\begin{itemize}
    \item This paper offers a theoretical interpretation of contrastive learning from the energy model's perspective and demonstrates its weakness to data contamination.

    \item We propose a unified model that applies a carefully designed energy function on unsupervised embedding learning and fine-tunes the embedding in a semi-supervised setting.
    
    \item Our model, \model{}, achieves the best or comparable detection performance over multiple benchmark datasets, despite leveraging only a small set of labeled samples.
    
    \item \model{} is robust to challenging  settings, where the training set contains unknown anomalies (Scenario-2 in experiments). Such property is important in real-world environments. \looseness=-1

\end{itemize}


%% file: 2_related_works.tex
\section{Related Works}
\noindent
Here we review studies on deep anomaly detection.

\cutparagraphup
\paragraph{Reconstruction-based learning} 
This approach recognizes the assumption that generative models cannot fluently recover data instances that are drawn from the out-of-distribution. The reconstruction error can be used as an anomaly score, where a high reconstruction error likely implies an anomaly. Recent studies proposed methods to define and utilize such reconstruction errors. For example, multiple autoencoders can define reconstruction error with a synthetically generated blurred image~\cite{choi2020novelty}. One study used reconstruction error along with latent vector to estimate novelty~\cite{abati2019latent}. Another study utilized gradients from back-propagation~\cite{kwon2020backpropagated}. However, studies have also found that anomalies do not always yield a high reconstruction error when classes are similar~\cite{zenati2018adversarially}. Likewise, it has been shown that the reconstruction error in low-level characteristics (i.e., pixel) cannot entail perceptual similarity~\cite{somepalli2021unsupervised}.

Some studies employed GAN (generative adversarial network) to complement the reconstruction loss, for instance, via utilizing a generator and a  discriminator~\cite{zenati2018adversarially} or using the reconstruction and discrimination loss~\cite{Perera_2019_CVPR,schlegl2017unsupervised}. A series of studies proposed exploiting  the generator's capability; for example, \cite{zaheer2020old} suggested training a network by distinguishing synthetic anomalies generated by an ensemble of the generator's old state with the current states of generator and discriminator. Alternatively, \cite{somepalli2021unsupervised} suggested the negative sampling method using the latent feature space distribution, where synthetic anomalies are sampled from the training data distribution with low probability. Some studies have regularized the latent feature space in GAN-based models~\cite{Perera_2019_CVPR, somepalli2021unsupervised}. However, it has also been reported that GAN-based models can produce a suboptimal solution and hence are inapplicable for complex datasets~\cite{liu2020energy,pang2020deep}.

\cutparagraphup
\paragraph{Self-supervised learning} This approach finds a representation by training a model with a surrogate task such as classification with pseudo-labels. Self-supervised learning approaches are known to produce a robust representation of normal data, which leads to low confidence in classification probability for anomaly detection. Transformations like rotation, shift, and patch re-arranging can be used to augment pseudo-labels~\cite{hendrycks2019selfsupervised, tack2020csi, wang2019effective}. As discussed earlier, the state-of-the-art method in this domain,  CSI~\cite{tack2020csi}, utilizes augmented pseudo-labels. The proposed model, \model{}, inherits the merit of self-supervised learning yet circumventing the possible weakness of data contamination, which is discussed later.


\cutparagraphup
\paragraph{One-class classifiers}
This approach tries to learn the decision boundary between the distribution of training data and OOD samples. \cite{ruff2018deep} was the first to suggest a loss function that forces training data samples to reside in a prefixed clustering centroid. The anomaly score is obtained by measuring the distance from each data point to the single centroid. \cite{Ruff2020Deep} extended the problem into a semi-supervised learning objective and set the loss function to pull all unlabeled or positive samples closer to the centroid while pushing negative samples away from the centroid. \cite{Bergman2020Classification-Based} improved the detection performance by separately assigning a centroid to each augmentation. However, these methods are limited in their representation ability if the data distribution is complex and heterogeneous. Our method, \model{}, follows the previous works' concepts while defining multiple prototypes as the centroids to solve the limitation and cover the data heterogeneity.  \looseness=-1

\cutparagraphup
\paragraph{Energy-based learning} Recently, OOD detection models using \textit{energy function} have been proposed~\cite{grathwohl2019your,liu2020energy}. The energy function measures the compatibility between a given input and a label~\cite{lecun2006tutorial}. This score can be used to detect anomalies since the energy function outputs low values for normal samples and high values for abnormal OOD samples. The logits obtained from the model have been used to define the energy function, often defined as the LogSumExp of the logits (see Eq.~\ref{eq:energy}). Existing works exploited an energy-based model on top of a standard discriminative classifier. \cite{grathwohl2019your} demonstrated that energy-based training of the joint distribution improves OOD detection. \cite{liu2020energy} proposed energy scores to distinguish in- and OOD samples, showing that this score outperforms the softmax confidence score. We also employ the insight of energy-based learning and its strength for OOD detection. \model{} defines the energy function derived from unsupervised CL objectives, which is designed to be robust against data contamination.


  

%% file: 3_background.tex
\vspace{-5mm}
\section{Background}
\vspace{-2mm}
\noindent
This section provides the necessary background on two relevant research lines: contrastive learning and energy-based model. Then, we offer a new insight that contrastive learning's objective does not align well with anomaly detection under data contamination. 

\cutsubsectionup
\subsection{Contrastive learning (CL)}
\vspace{-2mm}
\noindent
The core concept of CL is to train an encoder $f$ by maximizing agreement among similar images (i.e., positive samples) while minimizing agreement among dissimilar images (i.e., negative samples). Let $\mathbf{x}$ be an input query, and a set of positive and negative samples of $\mathbf{x}$ be denoted $\mathcal{X}_{+}$ and $\mathcal{X}_{-}$. The contrastive loss is defined as:
\begin{align}
L_{c}(\mathbf{x}) = -{1 \over |\mathcal{X}_{+}|} \log {{\sum_{\mathbf{x}' \in \mathcal{X}_{+}} \text{exp}({\text{sim}(f(\mathbf{x}), f(\mathbf{x}')) / \tau}})  \over {\sum_{\mathbf{x}' \in (\mathcal{X}_{+} \cup \mathcal{X}_{-})} \text{exp}(\text{sim}(f(\mathbf{x}), f(\mathbf{x}')) / \tau)}}, \label{eq:contrastive_loss}
\end{align}
where $\tau$ is the temperature value that controls entropy~\cite{hinton2015distilling} and sim($\cdot$) is the function that computes the similarity between two instances over the latent space.

We decompose the contrastive loss into two terms in Eq.~\ref{eq:simclr_decompose} as in~\cite{wang2020understanding}. First is the alignment loss ($L_{\text{align}}$), which encourages embeddings of positive samples to be closely positioned. Next is the uniformity loss ($L_{\text{uniform}}$), which matches all samples (whether positive or negative) into the pre-defined prior distribution with high entropy by pushing one another far away. Take cosine similarity, for example, as sim($\cdot$) as in SimCLR~\cite{chen2020simple}.
Then, the uniformity loss becomes the smallest when embeddings from both the positive and negative samples are normally distributed over the L2-normalized hypersphere.
\begin{align}
L_c(\mathbf{x}) = &\overbrace{-{1 \over |\mathcal{X}_{+}|} \log{\sum_{\mathbf{x}' \in \mathcal{X}_{+}}{\exp{(\text{sim}(f(\mathbf{x}), f(\mathbf{x}')) / \tau})} \nonumber}}^{\textstyle L_{\text{align}}(\mathbf{x})} \\
&\overbrace{+ {1 \over |\mathcal{X}_{+}|} \log \sum_{\mathbf{x}' \in (\mathcal{X}_{+} \cup \mathcal{X}_{-})} \text{exp}(\text{sim}(f(\mathbf{x}), f(\mathbf{x}')) / \tau)}^{\textstyle L_{\text{uniform}}(\mathbf{x})} \label{eq:simclr_decompose}
\end{align}
\begin{figure*}[t!]
\centering
\begin{subfigure}[t]{0.32\textwidth}
       \centering\includegraphics[height=3.6cm]{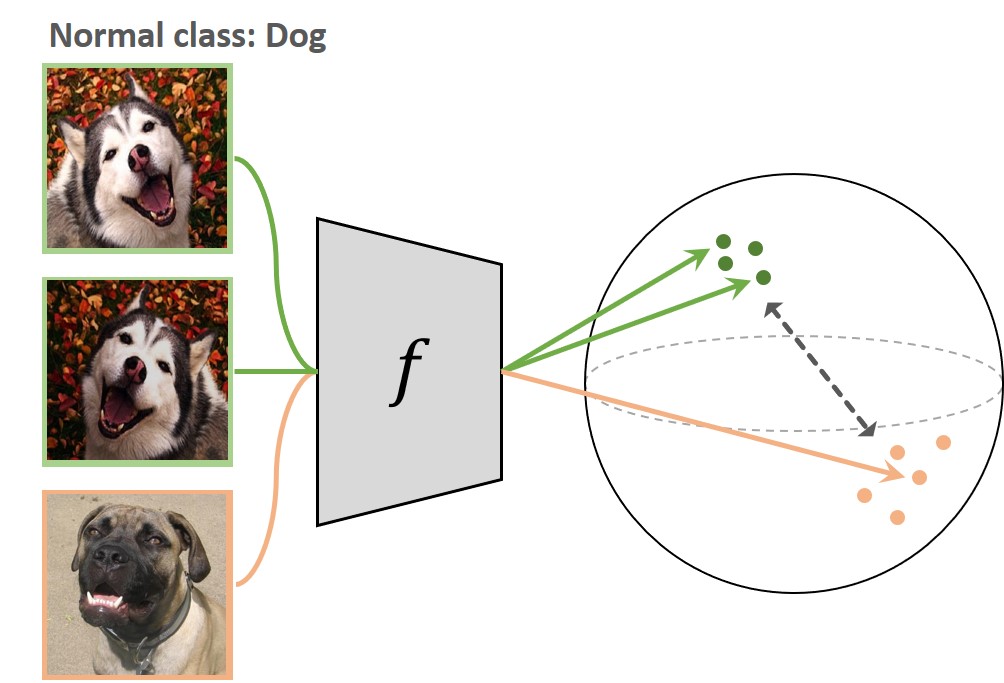}
      \caption{Step 1: Unsupervised CL for pre-training}
\end{subfigure}
\hspace{2mm}
\begin{subfigure}[t]{0.32\textwidth}
       \centering\includegraphics[height=3.6cm]{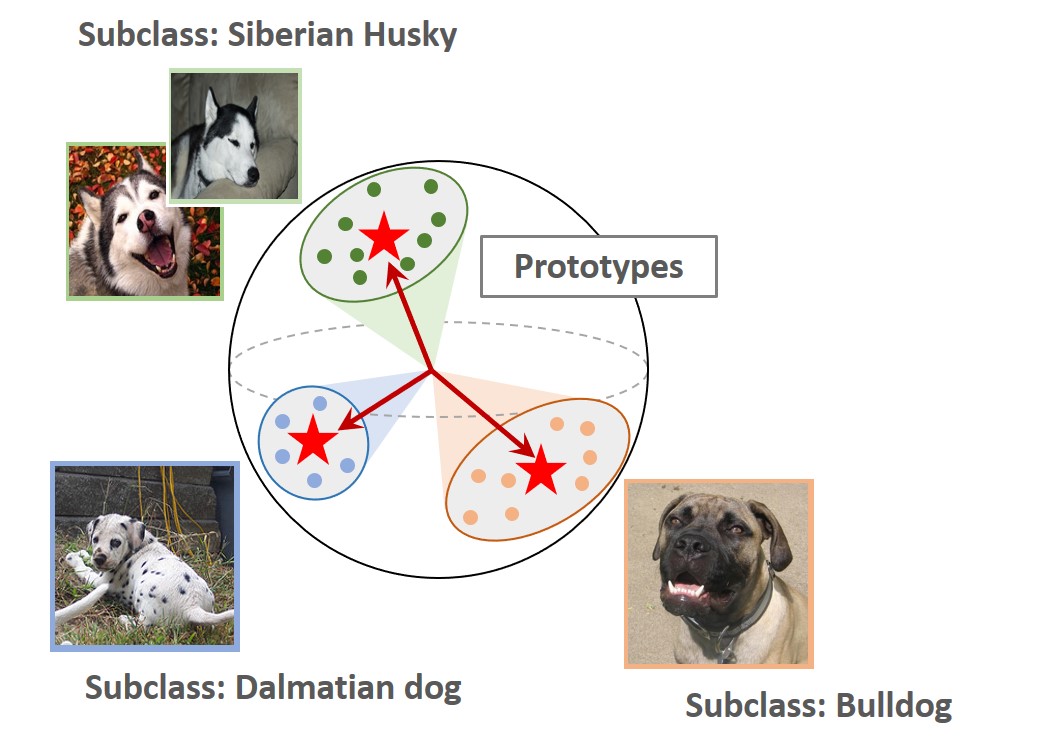}
      \caption{Step 2: Choosing prototypes}
      \label{fig:model_step2}
\end{subfigure}
\begin{subfigure}[t]{0.32\textwidth}
       \centering\includegraphics[height=3.6cm]{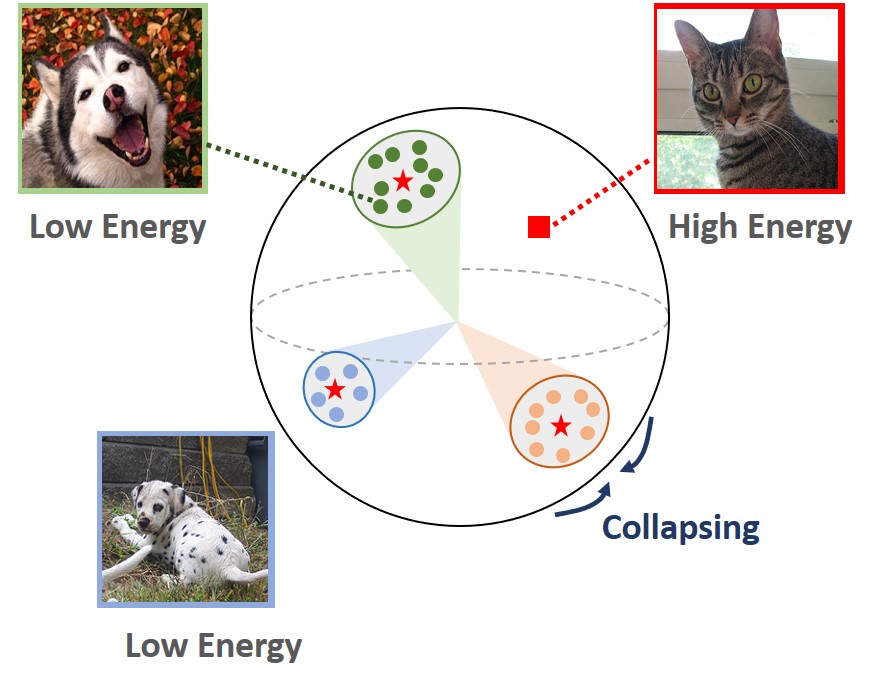}
      \caption{Step 3: Fine-tuning with prototypes}
\end{subfigure}
\caption{Illustration of \model{}. The model is pre-trained with the unsupervised contrastive learning objective during initialization. The model next chooses prototype vectors representing each subclass (e.g., different dog breeds). The model is then fine-tuned with the energy function derived from the similarity in embedding between the instance and the prototype vectors. The prototype vectors are periodically updated throughout training to fit in the fine-tuned data distribution.}
\label{fig:model}
\end{figure*}

\cutsubsectionup
\subsection{Energy-based model}

\noindent
Energy-based models~\cite{lecun2006tutorial} assume that any probability density function $p_\theta$ can be expressed as $E_\theta (\cdot)$:
\begin{align}
p_\theta(\mathbf{x}) = {\text{exp} (-E_\theta(\mathbf{x})) \over \int_{\mathbf{x}} \text{exp} (-E_\theta(\mathbf{x})) }.
\end{align}
The energy function $E_\theta (\cdot)$ maps each data point $\mathbf{x}$ to a scalar value that represents its fit to given data distribution. Since it is computationally intractable to calculate the normalizing factor $Z(\theta) = \int_{\mathbf{x}} \text{exp} (-E_\theta(\mathbf{x}))$, previous research has relied on approximations such as the Markov Chain Monte Carlo (MCMC) algorithm~\cite{brooks2011handbook,grathwohl2019your}. \looseness=-1

For the energy function choice, one can also consider a decision-making model with two variables $X$ and $Y$. In this scenario, the energy-based model defines the energy function $E_\theta (X, Y)$, and the model is trained to give low energy to a pair of samples of an image vector $\mathbf{x}$ and label $y$, $(\mathbf{x},y)$, that is drawn from the distribution of $(X,Y)$. The energy function can be transformed in the form of a conditional probability with temperature $\tau$ as in Eq.~\ref{eq:energy_fyx}~\cite{liu2020energy}. 
Then, $E_\theta (\mathbf{x})$ can be defined by marginalizing the energy function $E_\theta(\mathbf{x}, y')$ over $y'$, as in Eq.~\ref{eq:energy_fx}.
\begin{align}
    p_{\theta}(y | \mathbf{x}) &= {\text{exp} (-E_\theta (\mathbf{x}, y)/\tau)  \over \int_{y'} \text{exp}(-E_\theta (\mathbf{x}, y')/\tau) } = {\text{exp}(-E_\theta (\mathbf{x}, y)/\tau)  \over \text{exp}(-E_\theta(\mathbf{x})/\tau) } \label{eq:energy_fyx} \\
    &\ E_\theta (\mathbf{x}) = -\tau \cdot \text{log} \int_{y'} \text{exp}(-E_\theta (\mathbf{x}, y')/\tau) \label{eq:energy_fx}
\end{align}

\subsection{Rethinking the use of CL for anomaly detection}

\noindent
The objective of CL can be theoretically interpreted from the perspective of the energy function. For this, a discriminative classifier $h_\psi$ can be considered, which maps each data input to a logit vector and estimates the categorical distribution with the following softmax function:
\begin{align}
p_{\psi}(y | \mathbf{x}) = {\text{exp}(h_\psi (\mathbf{x})[y]) \over {\sum_{y'} \text{exp}({h_\psi (\mathbf{x})[y'])}}}, \label{eq:softmax}
\end{align}
where $h_\psi (\mathbf{x})[y]$ indicates the $y^{th}$ index of $h_\psi (\mathbf{x})$, i.e. the logit corresponding to the $y^{th}$ class label. Combining Eq.~\ref{eq:energy_fyx} and Eq.~\ref{eq:softmax}~leads a model to optimize with the following energy function $E_\psi (\mathbf{x})$. This marginalizes the $E_\psi (\mathbf{x}, y)$ over $y$~\cite{grathwohl2019your}. 
\begin{align}
E_\psi (\mathbf{x}) &= -\text{LogSumExp}_{y}(h_\psi (\mathbf{x})[y]) \nonumber \\
&= -\text{log}\sum_{y}\text{exp}(h_\psi (\mathbf{x})[y]) \label{eq:energy}
\end{align}

Similarly, we translate CL's objective as a classification task via considering each image instance to become a class of its own. Training a classifier by assigning the same pseudo-label to positive samples and a different label to negative samples will solve the objective of CL. If we denote $\hat{y}$ as a class label, the contrastive loss in Eq.~\ref{eq:contrastive_loss} can be re-defined in the form of the cross-entropy loss:
\begin{align}
L_c(\mathbf{x}) &= - {1 \over |\mathcal{X}_{+}|}\log{p(\hat{y} \in \mathcal{Y}_{+} | \mathbf{x})} \nonumber \\
p(\hat{y} \in \mathcal{Y}_{+} | \mathbf{x}) &= {{\sum_{\mathbf{x}' \in \mathcal{X}_{+}} \text{exp}({\text{sim}(f(\mathbf{x}), f(\mathbf{x}')) / \tau}})  \over {\sum_{\mathbf{x}' \in (\mathcal{X}_{+} \cup \mathcal{X}_{-})} \text{exp}({\text{sim}(f(\mathbf{x}), f(\mathbf{x}')) / \tau}})}, 
\label{eq:contrastive_cross-ent}
\end{align}
where $\mathcal{Y}_{+}$ is the set of pseudo-labels corresponding to $\mathcal{X}_{+}$ (i.e., positive samples of $\mathbf{x}$). Since the re-defined contrastive loss has the same form as in Eq.~\ref{eq:softmax}, we obtain the energy function by marginalizing $E (\mathbf{x}, \hat{y})$ over $\hat{y}$ in Eq.~\ref{eq:energy_cont}. Intuitively, this energy function represents how far an instance is placed from every training sample.
\begin{align}
E (\mathbf{x}) &= - \text{log} \sum_{\mathbf{x}' \in (\mathcal{X}_{+} \cup \mathcal{X}_{-})} \text{exp}({\text{sim}(f(\mathbf{x}), f(\mathbf{x}')) / \tau}) \nonumber \\
&\propto -L_{\text{uniform}}(\mathbf{x})
\label{eq:energy_cont}
\end{align}

According to Eq.~\ref{eq:simclr_decompose} and Eq.~\ref{eq:energy_cont}, the energy function from the contrastive objective should be negatively proportional to the uniformity loss ($L_{\text{uniform}}$). It means minimizing the contrastive loss leads to minimizing the uniformity loss and maximizing the given training samples' energy scores. This observation contradicts the original definition of the energy function, where a high energy value should correspond to the incompatible data configurations or anomalies. \looseness=-1 


%% file: 4_model.tex
\section{Energy-based Learning for Semi-supervised Anomaly Detection (\model{})}


\subsection{Model Description}

\noindent
We tackle the energy maximization problem via a fine-tuning step that combines an energy function with unsupervised contrastive pre-training. Figure~\ref{fig:model} illustrates these steps with an example. We also present an extension of this model, called \model{}+, that adopts several novel anomaly detection techniques. 

\cutparagraphup
\paragraph{\underline{Problem statement:} }
Let $\mathcal{D} = \{\mathbf{x}_i\}_{i=1}^{N}$ denotes a set of training images $\mathbf{x}_{i}$. In a semi-supervised problem setting, $\mathcal{D}$ can be divided into three disjoint sets $\mathcal{D} = \mathcal{X}_u \cup \mathcal{X}_n \cup \mathcal{X}_a$, where each stands for a set of unlabeled samples, labeled normal samples, and labeled anomaly samples. The main objective of anomaly detection is to train a normality score function $S(\mathbf{x})$ that represents the likelihood of a given instance $\mathbf{x}$ sampled from the normal data distribution. We assume that the majority of unlabeled samples $\mathcal{X}_u$ are normal, and thus let the model learn the (mostly) normal data distribution from $\mathcal{X}_u \cup \mathcal{X}_n$, while distinguishing the labeled anomaly set $\mathcal{X}_a$. \model{} involves the following three steps:

\cutparagraphup
\paragraph{(Step-1) Unsupervised contrastive pre-training}\mbox{}\\
\noindent
The first step initializes the encoder $f$ to learn general features from the normal data distribution. This step only uses the unlabeled set $\mathcal{X}_{u}$ and the labeled normal set $\mathcal{X}_n$, and pre-trains the encoder $f$ with an unsupervised CL approach, such as SimCLR~\cite{chen2020simple}.  Let $\hat{\mathbf{x}}^{(1)}$ and $\hat{\mathbf{x}}^{(2)}$ be two independent views of $\mathbf{x}$ from a pre-defined augmentation family $\mathcal{T}_a$ (i.e., $\hat{\mathbf{x}}^{(1)}$ = $t_{1}(\mathbf{x})$, $\hat{\mathbf{x}}^{(2)}$ = $t_{2}(\mathbf{x})$ where $t_{1}, t_{2} \sim \mathcal{T}_a$). Formally, the unsupervised CL loss on the given pair of images is defined as follows:
\begin{align}
L_{\text{CL}}(\hat{\mathbf{x}}^{(1)}, \hat{\mathbf{x}}^{(2)}) = - \log {{ \text{exp}({\text{sim}(f(\hat{\mathbf{x}}^{(1)}), f(\hat{\mathbf{x}}^{(2)})) / \tau}})  \over {\smashoperator{\sum\limits_{\mathbf{x}' \in \mathcal{\hat{B}}^{(-)}}} \text{exp}(\text{sim}(f(\hat{\mathbf{x}}^{(1)}), f(\mathbf{x}')) / \tau)}}, \label{eq:simclr_loss}
\end{align}
where $\mathcal{\hat{B}}^{(1)} = \{\hat{\mathbf{x}}_{i}^{(1)}\}_{i=1}^{m}$, $\mathcal{\hat{B}}^{(2)} = \{\hat{\mathbf{x}}_{i}^{(2)}\}_{i=1}^{m}$ denotes the batches with batch size $m$, and $\mathcal{\hat{B}}^{(-)} = \mathcal{\hat{B}}^{(1)} \cup \mathcal{\hat{B}}^{(2)} \setminus \hat{\mathbf{x}}^{(1)}$. $\tau$ and $\text{sim}(\cdot)$ are defined as before.

By assigning the augmented variations of each training data instance in $\mathcal{X}_u \cup \mathcal{X}_n$ as positive samples, the model can maximize agreement among those samples (i.e., the numerator in Eq.~\ref{eq:simclr_loss}). Simultaneously, all other instances in the same batch are treated as negative samples and pushed far away from one another in the latent space (i.e., the denominator in Eq.~\ref{eq:simclr_loss}). \looseness=-1


\cutparagraphup
\paragraph{(Step-2) Prototypes selection}\mbox{}\\ 
\noindent
Conventional unsupervised CL maximizes the energy scores of all training samples and distinguishes every instance in the latent space. To mitigate this issue, one may try to minimize the energy score of normal samples, which is identical to maximizing the uniformity loss (Eq.~\ref{eq:energy_cont}). This approach, however, will pull every embedded point into a single position. Instead, we propose a new energy function by assigning a pseudo-label $y_p$ to every training instance on the pre-trained embedding.


Let us define a set of prototypes $\mathcal{P}$ representing their subclasses in the normalized latent space. These prototypes can conceptually indicate heterogeneity of the training dataset. Then, every training sample should be mapped to the single nearest prototype $\mathbf{p} \in \mathcal{P}$ based on the cosine similarity. Following the form of the discriminative model, we may regard the encoder $f$ as a classifier that maps each data point to a prototype, where the categorical distribution of each pseudo-label is computed via the following softmax function: \looseness=-1
\begin{align}
p(y_p | \mathbf{x}) = {\text{exp}(\text{sim}(f(\mathbf{x}), \mathbf{p})) \over {\sum_{\mathbf{p}' \in \mathcal{P}} \text{exp}(\text{sim}(f(\mathbf{x}), \mathbf{p'}))}}. \label{eq:prototype_softmax}
\end{align}
This problem transformation lets us compute the energy function by marginalizing $E(\mathbf{x}, y_p)$ over $y_p$, similar to Eq.~\ref{eq:energy_cont}. The normality score function is then defined as a negation of the computed energy function (Eq.~\ref{eq:score_func}) with the temperature value $\tau$.
\begin{align}
S(\mathbf{x}) = - E(\mathbf{x}) = \text{log}\sum\limits_{\mathbf{p} \in \mathcal{P}}\text{exp}(\text{sim}(f(\mathbf{x}) , \mathbf{p}) / \tau) \label{eq:score_func}
\end{align}

To ensure prototypes are dispersed over the training data distribution, we choose centroids of each cluster to be prototypes. Spherical $k$-means clustering algorithm over the embeddings of training samples is considered in this work, with the cosine similarity as a distance metric.


\begin{table*}[t!]
\centering
\resizebox{1\textwidth}{!}{%
\begin{tabular}{c|ccccccccccc}
\hline
 $\gamma_{l}$ & OC-SVM~\cite{scholkopf2001estimating} & IF~\cite{liu2008isolation}   & KDE~\cite{ruff2018deep} & DeepSVDD~\cite{ruff2018deep}  & GOAD~\cite{Bergman2020Classification-Based} & CSI~\cite{tack2020csi}             & SS-DGM~\cite{kingma2014semi} & SSAD~\cite{gornitz2013toward}  & DeepSAD~\cite{Ruff2020Deep} & \model{} &\model{}+    \\ \hline
  .00         & 62.0   & 60.0 & 59.9  & 60.9      & 88.2 & \textbf{94.3}   &  -     & 62.0  & 60.9    &  -    &  -       \\
  .01         &        &      &      &           &      &                 & 49.7   & 73.0  & 72.6    &  80.0  &\textbf{94.3}\\ 
  .05         &        &      &       &           &      &                 & 50.8   & 71.5  & 77.9    & 85.7  &\textbf{95.2}\\
  .10         &        &      &       &           &      &                 & 52.0   & 70.1  & 79.8    & 87.1  &\textbf{95.5}\\ \hline
\end{tabular}}
\vspace{-3mm}
\caption{Experiment results on anomaly detection Scenario-1 over CIFAR-10. We change the ratio of labeled normal and anomalies in the training set and evaluate the model with averaged AUROC score over 90 experiments. Unsupervised algorithms are evaluated only for the settings with a labeling ratio $\gamma_t = .00$. Our extended model, \model{}+, achieves the best performance in all contaminated settings.}
\label{table:s1_ssl}
\end{table*}

\begin{table*}[t!]
\centering
\resizebox{1\textwidth}{!}{%
\centering
\begin{tabular}{c|ccccccccccc}
\hline
 $\gamma_{p}$ &  OC-SVM~\cite{scholkopf2001estimating} & IF~\cite{liu2008isolation}   & KDE~\cite{ruff2018deep} & DeepSVDD~\cite{ruff2018deep}  & GOAD~\cite{Bergman2020Classification-Based} & CSI~\cite{tack2020csi}             & SS-DGM~\cite{kingma2014semi} & SSAD~\cite{gornitz2013toward}  & DeepSAD~\cite{Ruff2020Deep} & \model{} &\model{}+     \\\hline
  .00         & 62.0   & 60.0 & 59.9  & 60.9     & 88.2 & 94.3  & 50.8    & 73.8  & 77.9    & 85.7   & \textbf{95.2} \\
  .05         & 61.4   & 59.6 & 58.1  & 59.6     & 85.2 & 88.2  & 50.1    & 71.5  & 74.0    & 83.5   & \textbf{93.0} \\
  .10         & 60.8   & 58.8 & 57.3  & 58.6     & 83.0 & 84.5  & 50.5    & 69.8  & 71.8    & 81.6   & \textbf{91.1} \\\hline
\end{tabular}}
\vspace{-3mm}
\caption{Experiment results on anomaly detection Scenario-2 over CIFAR-10. We change the contamination ratio in the training set and evaluate the model with averaged AUROC score over 90 experiments. For semi-supervised learning models, we fix the labeling ratio $\gamma_{l}$ to 0.05. Our extended model, \model{}+, achieves the best performance in all experiments.}
\label{table:s2_polluted}
\end{table*}

\cutparagraphup
\paragraph{(Step-3) Fine-tuning with prototypes}\mbox{}\\
\noindent
Finally, the model is fine-tuned via the following loss ($L_e$):
\begin{align}
L_e = \sum\limits_{\mathbf{x} \in \mathcal{X}_{a}}\frac{1}{C-S(\mathbf{x})}+\sum\limits_{\mathbf{x} \in \mathcal{X}_u\cup\mathcal{X}_n}\frac{1}{S(\mathbf{x})}, \label{eq:le}
\end{align}
where $C$ is a constant. This loss minimizes the normality scores of abnormal samples $\mathcal{X}_a$ and maximizes the scores of the mostly normal samples $\mathcal{X}_u \cup \mathcal{X}_n$. To stabilize this training process, we use an inverse form as the learning objective. According to the gradient of $L_e$ with respect to $\mathbf{x}$ (Eq.~\ref{eq:gradient_analysis}), the inverse of the quadratic term is multiplied on the gradient of the score function, $\nabla_{\mathbf{x}} S(\mathbf{x})$. This multiplier helps reduce the gradient signal when the score becomes sufficiently small for $\mathcal{X}_a$ or large for $\mathcal{X}_u \cup \mathcal{X}_n$. To ensure denominators remain positive, we set the constant $C$ as the largest possible value; given the input instance has the maximum similarity with all prototypes (i.e., $\text{sim}(f(\mathbf{x}), \mathbf{p})=1$ for all prototypes). \looseness=-1
\begin{align}
\nabla_{\mathbf{x}} L_e = &\sum\limits_{\mathbf{x} \in \mathcal{X}_a} \nabla_{\mathbf{x}} S(\mathbf{x})\Big(\frac{1}{C-S(\mathbf{x})}\Big)^2 \nonumber \\  &-\sum\limits_{\mathbf{x} \in \mathcal{X}_u \cup \mathcal{X}_n } \nabla_{\mathbf{x}} S(\mathbf{x})\Big(\frac{1}{S(\mathbf{x})}\Big)^2 
\label{eq:gradient_analysis}
\end{align}

Since we embed the data instances in $\mathcal{X}_u \cup \mathcal{X}_n$ nearby the chosen prototypes in the latent space, this process can be interpreted as the minimum volume estimation~\cite{scott2006learning} over the normalized latent space. It is connected to two works, Deep-SVDD~\cite{ruff2018deep} and Deep-SAD~\cite{Ruff2020Deep}, where the training instances are collapsed into a single centroid in the latent space. However, we utilize \textit{multiple} centroids (called `prototypes' in this research) to account for heterogeneity in data. Multiple centroids are better suited to learning distinct features from heterogeneous data; the same learning capability is hard for a single centroid. For example, Bulldogs and Siberian Huskies have distinctive visual features, yet they belong to the same class of dogs. Our model will assign these two dog types to different prototypes as in Fig.~\ref{fig:model_step2}.

As fine-tuning continuously changes the distribution of data instances in the latent space, one can no longer ensure the previous step's prototypes to be valid. Thus, we update the prototypes to fit the fine-tuned distribution periodically. This is done by repeating step-2 every few epochs in step-3. 

We also introduce the \textit{early stopping strategy} to avoid overfitting and guide the model to determine when to stop. The idea here is that the detection performance between the original images and their strongly augmented versions can be used as a validation indicator. Strong augmentations such as AutoContrast, Shear, and Cutout\cite{devries2017improved} can be regarded as tentative anomalies due to massive content-wise distortions~\cite{cubuk2020randaugment,sohn2020fixmatch}. We adopted the RandAugment algorithm~\cite{cubuk2020randaugment} for the choice of strong augmentation and separated the validation set from the unlabeled training set with a ratio of 5-to-95. We obtained the AUROC score between the original and augmented images to determine the final model that achieves the best AUROC score (see Fig.~\ref{fig:prototypes_scores}), denoted as the \textit{earlystop} score. \looseness = -1

\subsection{Extension with contrasting shifted instances}

\noindent
We extend \model{} in two ways. First, as suggested in~\cite{tack2020csi}, the training instances can be augmented via transformations called \textit{contrasting shifted instances}. These special rotations enlarge the training data size, help learn features more effectively, and enlarge the number of prototypes (i.e., allowing data heterogeneity). We augmented the training set by shifting transformations $\mathcal{T}_s$, and the augmented training set was used in both pre-training and fine-tuning. To ensure the transformed samples be assigned to different prototypes (even if they are from the same original image), we appended the cross-entropy loss $L_s$ throughout the training phase to predict the pseudo-label $y_t$ that represents which shifting transformation had been applied (Eq.~\ref{eq:multi_task}). \looseness=-1
\begin{align}
L_s = \sum_{\mathbf{x} \in \mathcal{X}_u \cup \mathcal{X}_n} \sum_{t_{s} \in \mathcal{T}_s} -\log p(y_t = t_{s} | t_{s}(\mathbf{x})) \label{eq:multi_task}
\end{align}

Second, employing an ensemble technique enhances the model's stability~\cite{opitz1999popular,ovadia2019can}. We iteratively calculated the normality scores from multiple views of the same image via random weak augmentation. Then, the ensembled score was computed by averaging the normality scores. More formally, given the set of augmentation family $\mathcal{T}_a$ and the set of shifting transformations $\mathcal{T}_s$, the ensembled normality score can be defined as in Eq.~\ref{eq:ensemble}. The ensembled score $S_{en}(\mathbf{x})$ is utilized to detect anomalies during evaluation. 
\begin{align}
 S_{en}(\mathbf{x}) =  \mathbb{E}_{t_a\sim \mathcal{T}_a, t_s \sim \mathcal{T}_s }[S( t_a\circ t_s (\mathbf{x}))] \label{eq:ensemble}
\end{align}

Implementing the above techniques -- contrasting shifted instances and ensemble technique -- on \model{}, we propose an extended model \model{}+. Algorithm details are described in the supplementary material.





%% file: 5_experiments.tex
\section{Experiments}
\subsection{OOD detection result}

\noindent
For evaluation, we consider three representative scenarios proposed from existing works on the CIFAR-10 dataset. The details of each scenario are described below.\footnote{Training details are reported in the supplementary material} \looseness=-1

\cutparagraphup
\paragraph{(Scenario-1) 
Semi-supervised 
classification~\cite{akcay2018ganomaly,Ruff2020Deep}.} 
Here we assume having access to a small subset of labeled normal $\mathcal{X}_n$ and anomalies $\mathcal{X}_a$ during training. One of the data classes in CIFAR-10 is set as in-distribution and let the remaining nine classes represent an anomaly. This means to sample $\mathcal{X}_a$ from the nine anomaly classes. Let the ratio of $\mathcal{X}_n$ and $\mathcal{X}_a$ both be denoted as $\gamma_{l}$. We then report the averaged AUROC scores on the test set over 90 experiments (10 normal $\times$ 9 anomaly) for a given $\gamma_{l}$. \looseness=-1

\begin{table}[t!]
\centering
\resizebox{.48\textwidth}{!}{%
\begin{tabular}{cl|cccc} 
\toprule
\multicolumn{2}{c|}{Datasets}            & GOAD & CSI  & \model{}+ \\ \midrule
                       & SVHN~\cite{svhn}            & 96.3 & \textbf{99.8} & \textbf{99.4} \\
                       & LSUN~\cite{liang2017enhancinglsun}            & 89.3 & 97.5 & \textbf{99.9} \\
CIFAR-10~\cite{krizhevsky2009learning} $\rightarrow$ & LSUN (FIX)~\cite{tack2020csi}      & 78.8 & 90.3 & \textbf{95.0} \\
                       & ImageNet (FIX)~\cite{tack2020csi}  & 83.3 & 93.3 & \textbf{96.4} \\
                       & CIFAR-100~\cite{krizhevsky2009learning}       & 77.2 & \textbf{89.2} & 86.3 \\ \bottomrule   
\end{tabular}}
\caption{Experiment results on anomaly detection in Scenario-3 over multiple datasets. Models are trained on CIFAR-10 and tested with the AUROC score. Compared to other baselines, \model{}+ shows the highest or comparable AUROC for detecting the outliers from various datasets.}
\label{table:s3_multi_data}
\end{table}


\cutparagraphup
\paragraph{(Scenario-2) Contaminated one-class classification~\cite{Ruff2020Deep}.} The next scenario tests the model's robustness under contamination in the training set, which is our utmost interest. It starts with the same setting as in Scenario-1. We assume the training data is polluted with a fixed ratio $\gamma_{p}$. This is done by sampling images from every anomaly class and adding them into the unlabeled set $\mathcal{X}_u$. We report the averaged AUROC scores over 90 experiments for each pollution ratio $\gamma_{p}$. The labeling ratio $\gamma_{l}$ is fixed to 0.05 for all experiments. \looseness=-1

\cutparagraphup
\paragraph{(Scenario-3) Auxiliary anomaly set~\cite{liu2020energy}.} This scenario tests whether the proposed model can leverage a large-scaled external dataset as an auxiliary anomaly set. We set all images in CIFAR-10 as in-distribution and let images from other datasets as anomalies. Then, we train the model with an auxiliary dataset (i.e., down-sampled ImageNet) as labeled anomalies and evaluate the detection performance on five other datasets with AUROC metrics.

\cutparagraphup
\paragraph{Results.} 
We discuss the results of each experiment. First, in the semi-supervised one-class classification in Scenario-1, \model{}+ achieves the state-of-the-art performance against all baselines (Table~\ref{table:s1_ssl}). The model works well even with a small set of labeled samples ($\gamma_l = 0.01$). The next experiment for Scenario-2 tests the model's performance against data contamination (Table~\ref{table:s2_polluted}), which shows both \model{} and \model{}+ to be stable against contamination in the training data. A significant performance drop seen for CSI supports our claim that the CL objective alone fails to handle data contamination. In contrast, the energy-based fine-tuning step can alleviate this problem and achieve outstanding performance. Lastly, Table~\ref{table:s3_multi_data} shows the result for Scenario-3 on the model's ability to leverage the external dataset as an auxiliary outlier. Compared to other strong baselines, \model{}+ shows the highest or comparable detection performance for all datasets. Results on the other benchmark dataset are reported in the supplementary materials. \looseness=-1

\subsection{Component analyses}
\noindent
We also test the contribution of each component in \model{}+ in three critical analyses. We regard the plane class in CIFAR-10 as the normal data and fix the labeling ratio $\gamma_l$ and pollution ratio $\gamma_p$ to 0.05 (Scenario-2) for the remainder of this section. \looseness=-1

\begin{table}[t!]
\centering
\resizebox{.48\textwidth}{!}{%
\begin{tabular}{ll|c}
\toprule
\multicolumn{2}{l|}{Ablations}                          & AUROC (\%) \\ \hline
\multicolumn{1}{c|}{\multirow{2}{*}{Score function}} &  Cosine similarity: $S_{cos}(\mathbf{x})$  &  89.1     \\ 
\multicolumn{1}{c|}{}                                &  Energy from CL objective: $S_{cont}(\mathbf{x})$ & 81.6      \\ \midrule
\multicolumn{1}{c|}{\multirow{2}{*}{Loss objective}} &   Naive loss: $L_{naive}$  &   82.6   \\ 
\multicolumn{1}{c|}{}                                &  DeepSAD loss: $L_{sad}$ & 90.3 \\ \midrule
\multicolumn{2}{c|}{\model{}+ (Ours)}                        &  \textbf{91.4}   \\ \bottomrule
\end{tabular}}
\caption{Ablation study results on the score function and loss objective over CIFAR-10. Both label ratio $\gamma_{l}$ and pollution ratio $\gamma_{p}$ are set to 0.05. Our loss objective with the energy score function recorded achieves the best performance.}
\label{table:loss_and_score}
\end{table}

\cutparagraphup
\paragraph{Ablation study on the score and loss function.} \mbox{} \\
We explore several possible score functions and objectives as alternatives to the proposed ones, thereby measuring each component's contribution. The description of each ablation is described below. The first two are ablations on alternative score functions. \smallskip


\begin{itemize}
    \item \underline{Cosine similarity.}
    A normality score can be obtained by measuring the cosine similarity of the given sample and the nearest prototype vector. 
    \vspace*{-1mm}
    \begin{align}
    S_{cos}(\mathbf{x}) = max_{ \mathbf{p} \in \mathcal{P}}(f( \mathbf{x}),\mathbf{p}) 
    \label{eq:cosine_score}
    \end{align} 
    \item \underline{Energy from CL objective.}
    The uniformity loss in CL objective (Eq.~\ref{eq:energy_cont}) can be utilized as a normality score. With the given augmented batch $\mathcal{B}$ from the training set, we define the score function as Eq.~\ref{eq:contrastive_score} during training. For evaluation, instead of using batch $\mathcal{B}$ from the test set, the normality score is computed over the training set (i.e., $\mathcal{X}_n \cup \mathcal{X}_u$).
\end{itemize}
\vspace*{-1mm}
\begin{align}
    S_{cont}(\mathbf{x}) = \text{log} \sum\limits_{\mathbf{x}' \in \mathcal{B} \setminus \{\mathbf{x}\}} \text{exp}(\text{sim}(f(\mathbf{x}),f(\mathbf{x}')))
\label{eq:contrastive_score}
\end{align}

\noindent
The next two are ablation on alternative loss objectives:
\begin{itemize}
    \item \underline{Naive loss}. The naive loss is the simplest form that maximizes the score for normal samples as a negative form of score function ($-S(\mathbf{x}))$ while minimizing for anomalies as a positive form of score function ($S(\mathbf{x}))$.
    \vspace*{-1mm}
    \begin{align}
        L_{naive} = \sum_{\mathbf{x} \in \mathcal{X}_a}{S(\mathbf{x})} + \sum\limits_{\mathbf{x} \in \mathcal{X}_n \cup \mathcal{X}_u}{-S(\mathbf{x})}
        \label{eq:loss_naive}
    \end{align}
    \item \underline{DeepSAD loss}. This form of loss is introduced in DeepSAD~\cite{Ruff2020Deep}. The score for anomalous samples are maximized as the inverse form of the score function.
    \vspace*{-1mm}
    \begin{align}
        L_{sad} = \sum_{\mathbf{x} \in \mathcal{X}_a}{\frac{1}{C-S(\mathbf{x})}} + \sum\limits_{\mathbf{x} \in \mathcal{X}_n \cup \mathcal{X}_u}{-S(\mathbf{x})}
        \label{eq:deepsad_loss}
    \end{align}
\end{itemize}

Table~\ref{table:loss_and_score} shows the results, where all ablations lead to a substantial performance drop. Specifically, changing the CL objective's energy function led to the most extensive degradation, reinforcing our theoretical interpretation presented in \S3.3. Our loss objective design, which exploits the inverse form of the energy function for both anomaly and normal samples, achieves the best performance compared to all ablations. \looseness=-1  

\begin{figure}[t!]
\centering
\begin{subfigure}[t]{0.225\textwidth}
       \centering\includegraphics[height=2.95cm]{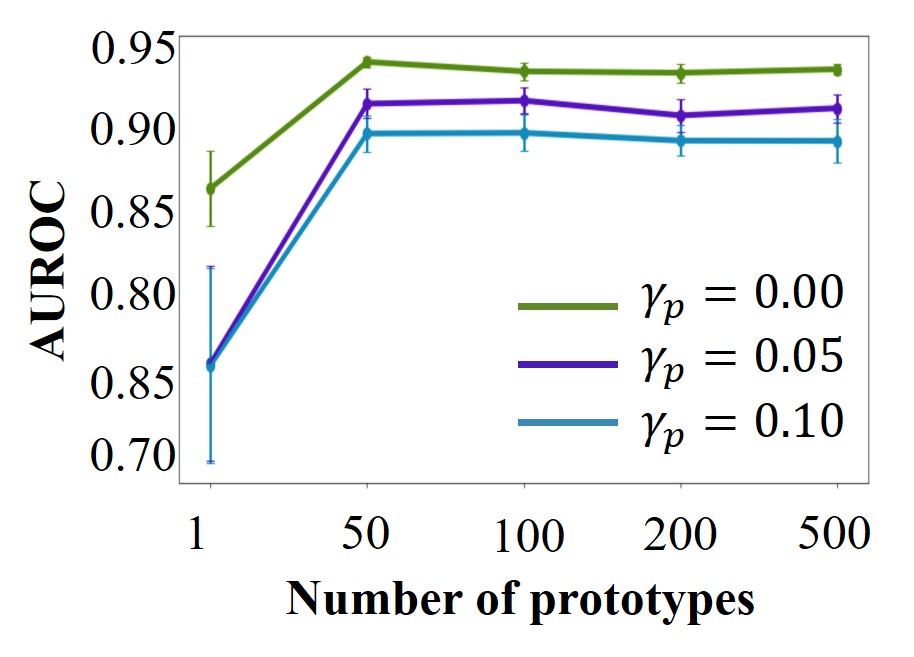}
       \caption{Performance with varying the number of prototypes}
       \label{fig:num_proto_score}
\end{subfigure}
\hspace{2mm}
\begin{subfigure}[t]{0.225\textwidth}
       \centering\includegraphics[height=2.9cm]{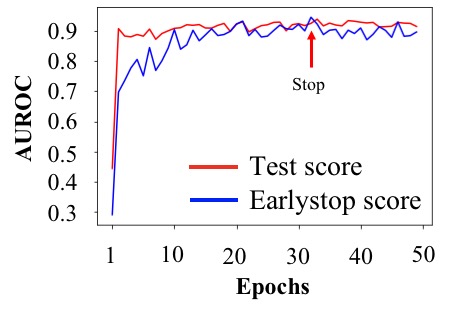}
       \caption{Classification performance and early-stop score}
       \label{fig:prototypes_scores}
\end{subfigure}
\caption{Analyses on the number of prototypes and early stopping. All experiments are done over CIFAR-10, where the plane class is normal, and other classes are anomalous. \model{}+ shows successful performance over different contamination ratios with a reasonably large number of prototypes. ($\gamma_p$ denotes the contamination ratio)}
\label{fig:analyses2}
\vspace{-3mm}
\end{figure}

\cutparagraphup
\paragraph{Analyses on the number of prototypes} \mbox{} \\
We investigate the dependency between the prototype count and \model{}+ performance. The prototype count decides the model's capacity for handling heterogeneous data types within the normal distribution and directly impacts the overall performance. Setting this number too small or too large will lead to underfitting or overfitting. We analyzed the prototype count's effect by varying it to 1, 50, 100, 200, and 500. Figure \ref{fig:num_proto_score} shows the mean AUROC score with standard error over three different contamination ratios ($\gamma_p$ = 0.00, 0.05, 0.10). For the count of 1, the model fails to converge, and the early stop strategy does not work in order. In contrast, for the large enough count, e.g., 100, the model converges with a successful result. Given the prototype count is set to a reasonably large value, \model{}+ consistently produces high-performance results. \looseness=-1  
\cutparagraphup
\paragraph{Analysis on the early stopping strategy} \mbox{} \\
The early stopping strategy is another important factor to be examined. The earlystop score is computed by the AUROC score of the task that distinguishes between the input image and its strongly augmented versions. This analysis reveals a highly positive correlation between the earlystop score and the actual model performance (Pearson correlation 0.912$\pm$0.038). This implies that the proposed earlystop score can represent the actual performance on the test set. Furthermore, Figure~\ref{fig:prototypes_scores} shows that the earlystop score eventually converges after some amount of time, and it gives an appropriate timing for earlystop with high detection performance on the test-set.

\begin{figure}[t!]
\centering
\begin{subfigure}[t]{0.22\textwidth}
       \centering\includegraphics[height=3.1cm]{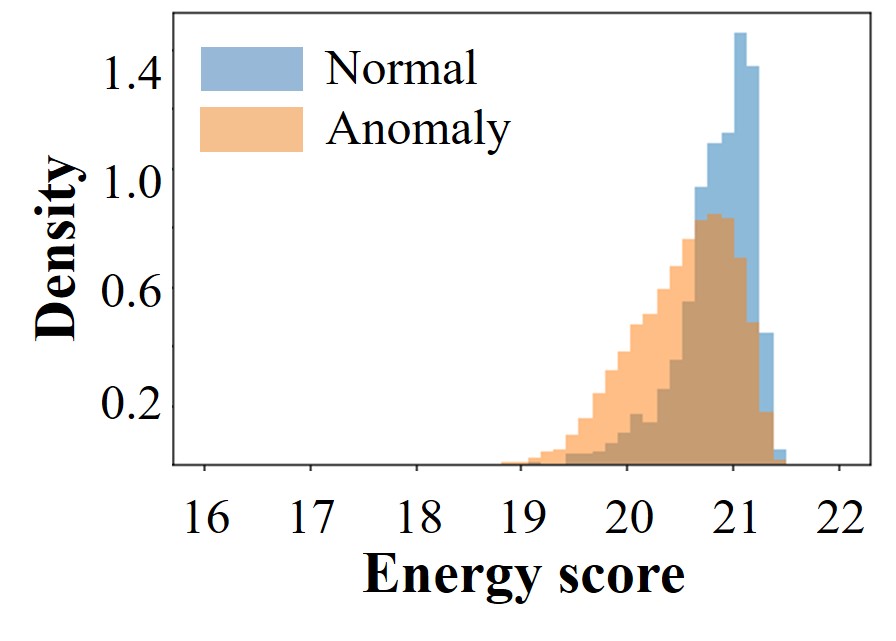}
\end{subfigure}
\hspace{3mm}
\begin{subfigure}[t]{0.22\textwidth}
       \centering\includegraphics[height=3.1cm]{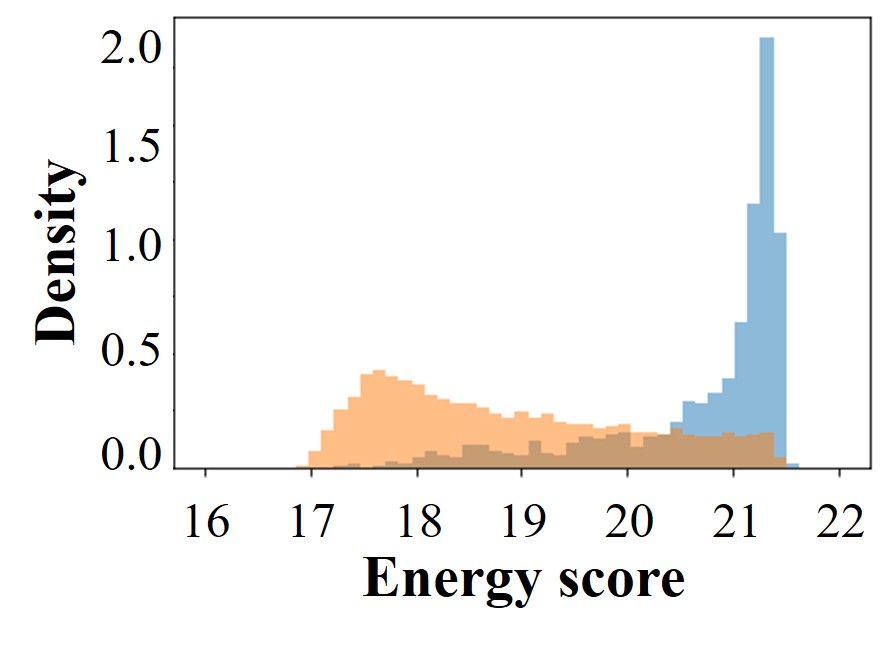}
\end{subfigure}
~~~~~~~~~~~~~~~~~~~~~~~~\ \ \ \ \ $~~~~~~~~~~~~$(a) Epoch 1~~~~~~~~~~~$~~$~~~~~~~~~~~~~~~~~~~(b) Epoch 2\\
\caption{Histogram of the distributions of scores across training epochs. The energy distribution of normal samples is sharpened while anomaly samples are spread out as the fine-tuning process goes on.}
\label{fig:energy_density}
\end{figure}

\begin{figure}[t!]
\centering\includegraphics[width=0.29
\textwidth]{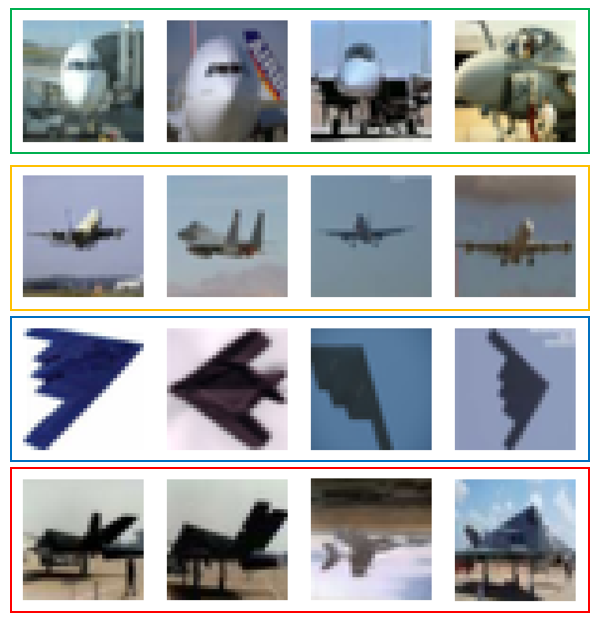}
\vspace{-2mm}
\caption{Example images sampled from each of prototype classes (Airplane class in CIFAR-10). Images in a row are from the same class, and they share a common shape that is distinguishable by human eyes.}
\vspace{-2mm}
\label{fig:prototype}
\end{figure}

\subsection{Qualitative analyses}
\noindent
We now examine the inner workings of \model{}+ in detail. Figure~\ref{fig:energy_density} shows the energy score distribution at epoch 1 and epoch 2, which shows separation to already begin in epoch 2. At around epoch 30, the two distributions are nearly separated (omitted due to space). The fine-tuning process of \model{}+ separates normal and anomaly samples so that the distribution of normal is sharpened nearby the prototypes and the distribution of anomaly is spread out. \looseness=-1

Figure~\ref{fig:prototype} shows examples of which training samples are positioned to the same prototypes after training. Images in the same row (i.e., belong to the same prototype) exhibit a common shape that is distinguishable in human eyes. These examples confirm that \model{}+'s flexible and superior embedding effectively explains the distribution of normal samples.


%% file: 6_discussion.tex
\section{Conclusion}
\noindent
We presented a unified energy-based approach for semi-supervised anomaly detection. We demonstrated that the contrastive learning objective is potentially fragile in the contamination scenarios interpreted from the energy-based perspective. We suggested a new energy function concerning prototypes and introduced the fine-tuning process in this light. With these components, the proposed model \model{} and \model{}+ could successfully distinguish anomalies from normal samples while leveraging the high representation power of unsupervised contrastive learning. \model{}+ achieves SOTA performance among other baselines and shows strong robustness against the contamination of unknown anomalies. We believe our effort will renew the interest in energy-based learning of OOD detection. \looseness=-1

%% file: 7_Appendix.tex
\appendix

\section{Implementation details}
We use ResNet-18 as a backbone network. \model{} is pre-trained with the loss objective $L_\text{CL}$ (Eq.~\ref{eq:simclr_loss}) for 500 epochs and fine-tuned with the loss objective $L_e$ (Eq.~\ref{eq:le}) for 50 epochs. The batch size for the pre-training and fine-tuning step is set to 512 and 64, respectively. $\tau$ is set to 0.5. For the optimization, we utilize the LARS optimizer with a learning rate of 0.1 and weight decay $10^{-6}$ for the pre-training, and utilize the Adam optimizer with a learning rate of $10^{-4}$ for the fine-tuning step. The number of prototypes is set to 50, and we update prototypes in \model{} every epoch.

In the case of \model{}+, we follow the architecture and training details of CSI~\cite{tack2020csi} for the entire pre-training step. Then, \model{}+ is fine-tuned with the loss objective $L_e + L_s$ (Eq.~\ref{eq:le},~\ref{eq:multi_task}) for 50 epochs. We adopt Adam optimizer with a learning rate of $10^{-4}$ for the optimization. The batch size is set to 64, and the number of sampling instances for the ensemble technique is set to 10. The number of prototypes is set to 100. We update prototypes for every three epochs.

For the augmentation family $\mathcal{T}_a$, we select the set of transformations from SimCLR~\cite{chen2020simple} in both the pre-training and fine-tuning process (see Table~\ref{Tab:SimCLR_aug}). We use rotation by multiples of 90 degrees (e.g., 90$^{\circ}$, 180$^{\circ}$, 270$^{\circ}$, 360$^{\circ}$) as shifting transformations $\mathcal{T}_s$. For the strong augmentation, the set of transformations in RandAugment~\cite{cubuk2020randaugment} except rotation is excerpted and utilized for the earlystop strategy (see Table~\ref{Tab:Strong_aug}). We set the number of transformations for RandAugment to 12 ($n=12$), transform magnitude to 5 ($m=5$), and modify the probability for applying the transformation from 0.5 to 0.8 to ensure providing anomalies. 

\begin{figure}[!htbp]
\centering
\begin{minipage}{0.48\textwidth}
\centering
\scalebox{0.9}{
\begin{tabular}{l|cc}
\toprule
Transformation & Parameter & Value \\ \midrule
ColorJitter    & B,C,s,h,p &  0.4, 0.4, 0.4, 0.1, 0.8\\
RandomGrayScale  & P & 0.2\\ 
RandomResizedCrop & w,h & 0.54, 0.54 \\ 
\bottomrule 
\end{tabular}}
\caption{List of transformations used for $\mathcal{T}_a$.}
\label{Tab:SimCLR_aug}
\end{minipage}
\hspace{1mm}
\begin{minipage}{0.48\textwidth}
\centering
\scalebox{1}{
\begin{tabular}{l|cc}
\toprule
Transformation & Parameter & Range \\ \midrule
AutoContrast    &  - &  -\\
Equalize    &  - & -\\
Identity    &  - & -\\
Brightness  &  $B$ & [0.01, 0.99]\\
Color&   $C$ & [0.01, 0.99]\\
Contrast    &  $C$ & [0.01, 0.99]\\
Posterize    &  $B$ & [1, 8]\\
Sharpness & $S$ & [0.01, 0.99]\\
Shear X, Y &  $R$ & [-0.3, 0.3]\\
Solarize &  $T$ & [0, 256]\\
Translate X, Y &  $\lambda$ & [-0.3, 0.3]\\
\bottomrule 
\end{tabular}}
\caption{List of transformations \model{}+ used for strong augmentation. Those are excerpted from RandAugment~\cite{cubuk2020randaugment}.}
\label{Tab:Strong_aug}
\end{minipage}
\end{figure}



\section{Algorithm description}
\noindent
In this section, we provide a PyTorch-style pseudo-code of earlystopping, training, and evaluation of \model{}+. 

\definecolor{dkgreen}{rgb}{0,0.6,0}
\definecolor{gray}{rgb}{0.5,0.5,0.5}
\definecolor{mauve}{rgb}{0.58,0,0.82}
\definecolor{green}{rgb}{0.8, 0.47, 0.196}
\lstset{frame=tb,
  aboveskip=3mm,
  belowskip=3mm,
  language=Python,
  showstringspaces=false,
  columns=flexible,
  basicstyle={\small\ttfamily},
  numbers=none,
  numberstyle=\tiny\color{gray},
  keywordstyle = {\color{green}},
  keywordstyle = [2]{\color{blue}},
  keywordstyle = [3]{\color{black}},
  keywordstyle = [4]{\color{teal}},
  otherkeywords = {earlystop_score, Concat, Rotate, Matmul, calculate_energy, roc_auc_score, Weak_aug, Strong_aug,CEloss,
  logits},
    morekeywords = [2]{earlystop_score,calculate_energy},
    morekeywords = [3]{logits,images},
    morekeywords = [4]{Concat, Rotate, Matmul, Weak_aug, Strong_aug, roc_auc_score,CEloss,Where,update 
    , Ones_like},
  commentstyle=\color{dkgreen},
  stringstyle=\color{mauve},
  breaklines=true,
  breakatwhitespace=true,
  tabsize=4
}
\begin{lstlisting}
"""
Earlystop score algorithm
"""

def earlystop_score(model, prototypes, valid_set):
    aucscores = []
    for images in valid_set:
    	batch_size = image.size(0)
    	images1, images2 = Weak_aug(images), Weak_aug(Strong_aug(images))
    	images1 = Concat([Rotate(images1, 90 * d) for d in range(4)])
    	images2 = Concat([Rotate(images2, 90 * d) for d in range(4)])
        label_list = [1] * batch_size + [0] * batch_size
        
		prob1 = calculate_energy(model, images1, prototypes)
    	prob2 = calculate_energy(model, images2, prototypes)
    	
        aucscores.append(roc_auc_score(label_list, prob1 + prob2))
    return mean(aucscores)

def calculate_energy(model, images, prototypes):
    z = model(images)
    logits = Matmul(z, prototypes.t())
    logits_list = logits.chunk(4, dim = 0)
    Px_mean = 0
    for shi in range(4):
        Px_mean += log(sum(exp(logits_list[shi])))
    return Px_mean

"""
Fine tune algorithm
T: Temperature
semi_targets: 0,1,-1 for unlabeled, labeled normal, labeled anomalies
"""
def train(model, train_set, prototypes)
	for images, semi_targets in train_set:
		images1, images2 = Weak_aug(images), Weak_aug(images) 
		images1 = Concat([Rotate(images1, 90 * d) for d in range(4)]) # 4B
    	images2 = Concat([Rotate(images2, 90 * d) for d in range(4)]) # 4B
		images_pair = Concat([images1, images2]) # 8B
		num_prototypes = prototypes.size(0) # P 

		shift_labels = Concat([Ones_like(semi_targets) * d for d in range(4)]) 
		semi_targets = semi_targets.repeat(8) # 8B

		z = model(images_pair) # 8B x D 
		logits = Matmul(z, prototypes.t()) / T # 8B x P
		
		C = log(num_prototypes + 1/T)
		scores = log(sum(exp(logits))) # 8B 

		Le = mean(Where(semi_targets == -1, (C - scores) ** -1, scores ** -1)) 
		Ls = CEloss(z, shift_labels)
		Loss = Le + Ls

		Loss.backward()
		update(model.params)

"""
Test algorithm
n_samples: number of sampling instances for ensembling
"""
def test(model, test_set, prototypes, n_samples)
	z = 0
	scores = []
	total_semi_targets = []

	# semi_targets: 0,1,-1 for unlabeled, labeled normal, labeled anomalies
	for images, semi_targets in test_set:
		for seed in range(n_samples): # ensembling
			images_aug = Weak_aug(images)
			images_aug = Concat([Rotate(images_aug, 90 * d) for d in range(4)])
			z += model(images_aug)
		z /= num_samples
		logits = Matmul(z, prototypes.t())
		logits_list = logits.chunk(4, dim = 0)

		Px_mean = 0
		for shi in range(4):
			Px_mean += log(sum(exp(logits_list[shi]))) 
		total_semi_targets.extend(semi_targets)
		scores.extend(Px_mean)
	auroc = roc_auc_score(total_semi_targets, scores)
	return auroc 
\end{lstlisting}

\section{Full results}

\subsection{OOD detection results -- CIFAR-10}
\input{tables/full_res_p00_l001.tex}
\input{tables/full_res_p00_l005.tex}
\input{tables/full_res_p00_l010.tex}

\input{tables/full_res_p005_l001.tex}
\input{tables/full_res_p005_l005.tex}
\input{tables/full_res_p005_l010.tex}

\input{tables/full_res_p010_l001.tex}
\input{tables/full_res_p010_l005.tex}
\input{tables/full_res_p010_l010.tex}

\subsection{OOD detection results -- ImageNet-10}
We evaluate \model{}+ on ImageNet-10, which is a 10-class subset of the original ImageNet dataset. The total number of ImageNet-10 samples is 13,000, and they are separated into train and test set with a ratio of 10-to-3. To adjust the prototype count according to the sample size (i.e., 1000 images per class), we set a prototype count to 20. Then, the batch size is set to 32, and an Adam optimizer with a learning rate of $5e^{-5}$ is utilized during the fine-tuning process. All other remaining parameter-settings are the same as settings from CIFAR-10 experiments. 

Table~\ref{tab:imagenet_nocontam} and Table~\ref{tab:imagenet_contam} describe the anomaly detection results over the ImageNet-10 dataset. For both scenarios, \model{}+ achieves comparable performance against CSI. Even though \model{}+ shows a small drop in the detection performance when it is trained on the clean training set, we find that \model{}+ is much robust to the contamination. 

\input{tables/cont_res0.0.tex}
\input{tables/cont_res_0.1}

\newpage

\subsection{Qualitative analysis}
\begin{figure}[H]
\centering
\begin{subfigure}[t]{0.3\textwidth}
       \centering\includegraphics[width=1\textwidth]{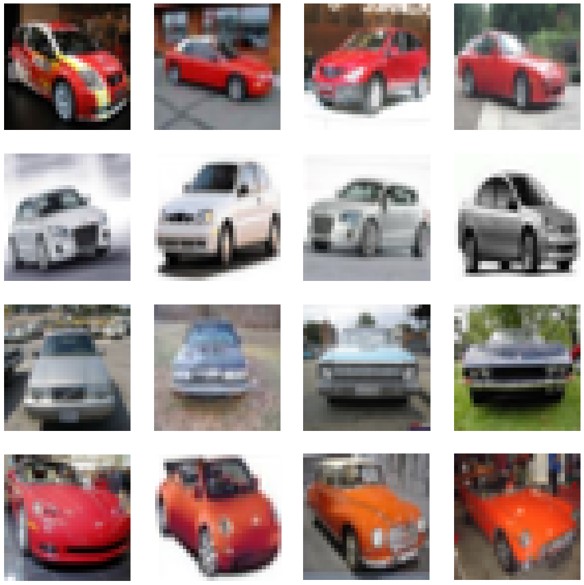}
       \caption{Car}
\end{subfigure}
\hspace{3mm}
\begin{subfigure}[t]{0.3\textwidth}
       \centering\includegraphics[width=1\textwidth]{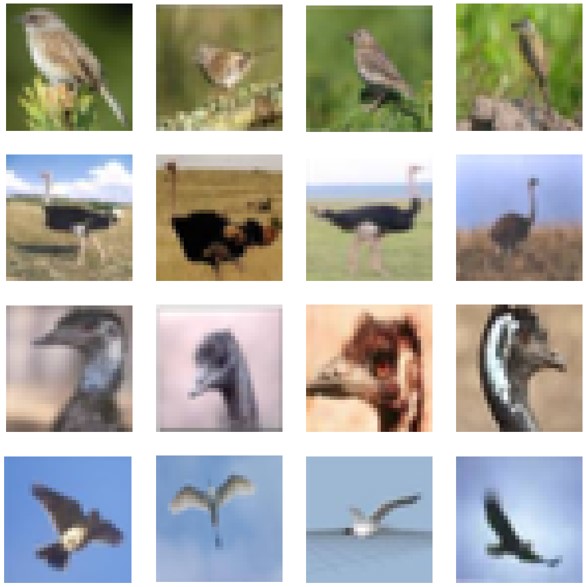}
       \caption{Bird}
\end{subfigure}
\hspace{3mm}
\begin{subfigure}[t]{0.3\textwidth}
       \centering\includegraphics[width=1\textwidth]{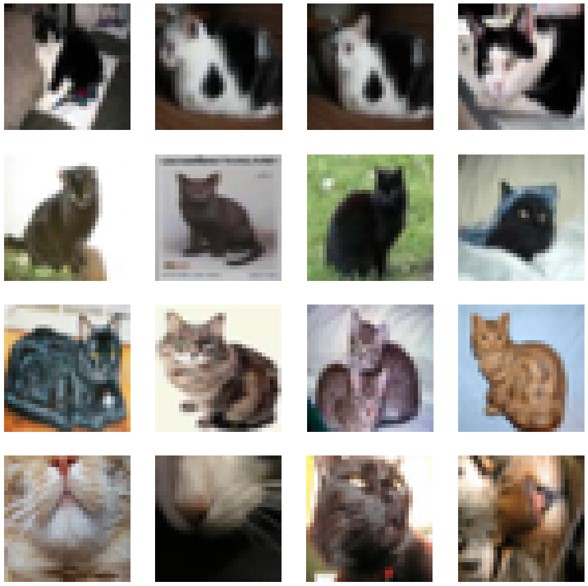}
       \caption{Cat}
\end{subfigure}
\begin{subfigure}[t]{0.3\textwidth}
       \centering\includegraphics[width=1\textwidth]{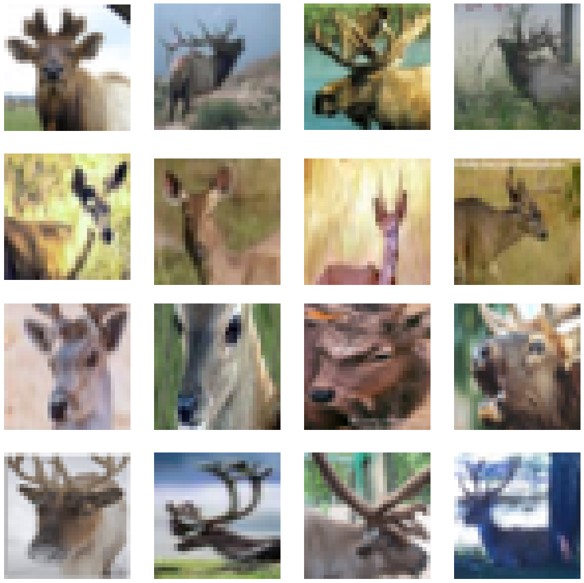}
       \caption{Deer}
\end{subfigure}
\hspace{3mm}
\begin{subfigure}[t]{0.3\textwidth}
       \centering\includegraphics[width=1\textwidth]{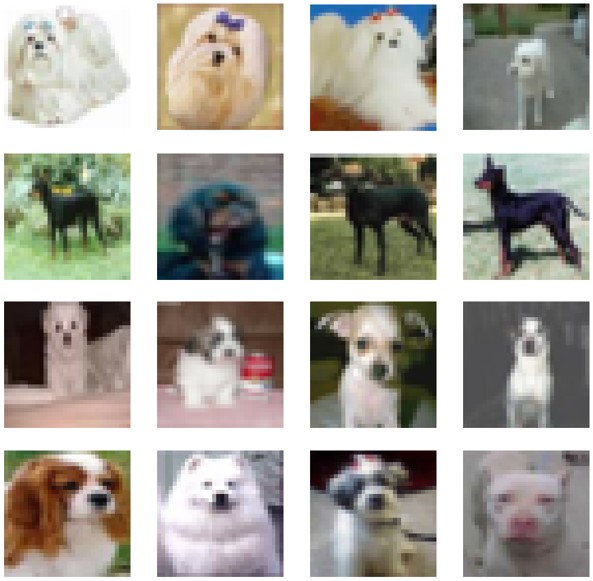}
       \caption{Dog}
\end{subfigure}
\hspace{3mm}
\begin{subfigure}[t]{0.3\textwidth}
       \centering\includegraphics[width=1\textwidth]{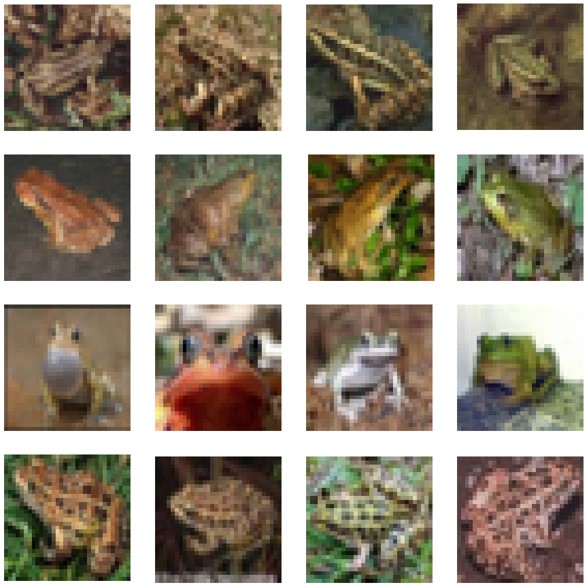}
       \caption{Frog}
\end{subfigure}
\begin{subfigure}[t]{0.3\textwidth}
       \centering\includegraphics[width=1\textwidth]{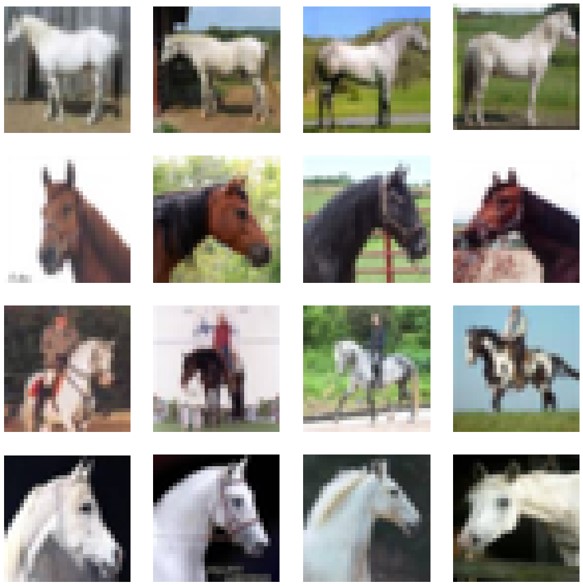}
       \caption{Horse}
\end{subfigure}
\hspace{3mm}
\begin{subfigure}[t]{0.3\textwidth}
       \centering\includegraphics[width=1\textwidth]{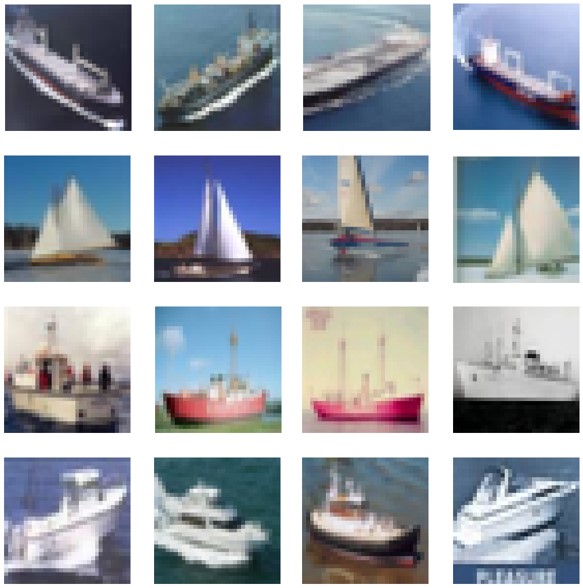}
       \caption{Ship}
\end{subfigure}
\hspace{3mm}
\begin{subfigure}[t]{0.3\textwidth}
       \centering\includegraphics[width=1\textwidth]{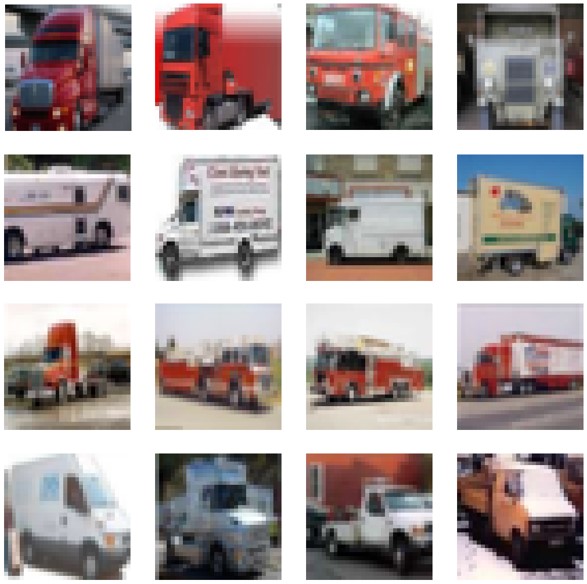}
       \caption{Truck}
\end{subfigure}
\caption{Example images sampled from each of prototype classes (class \#1 to \#9 in CIFAR-10). Images in a row are from the same prototype class.}
\label{fig:qualitative_full}
\end{figure}

%% file: tables/full_res_p00_l001.tex
\begin{table}[H]
\centering
\begin{minipage}{0.90\textwidth}
\centering
\begin{tabular}{ccccccccccc|c}
\hline
\multicolumn{1}{r}{} & \multicolumn{1}{r}{Plane} & Car & Bird & Cat & Deer & Dog & Frog & Horse & Ship & Truck & Mean \\ \hline
Plane & -  & 94.2 & 89.3 & 93.5 & 92.4 & 91.6 & 93.8 & 94.7 & 90.3 & 94.3 &    92.7 \\
Car   &  98.9 & - & 98.2 & 99.0 & 99.1 & 98.6 & 97.7 & 99.0 & 98.8 & 99.1 &    98.7 \\
Bird  &  88.9 & 92.0 & - & 93.5 & 91.9 & 91.9 & 83.8 & 92.6 & 91.4 & 91.7 &    90.9 \\
Cat   &  89.0 & 87.8 & 85.2 & - & 90.3 & 76.9 & 77.5 & 85.9 & 87.1 & 87.8 &    85.3 \\
Deer  &  95.5 & 95.5 & 94.9 & 96.0 & - & 93.3 & 92.3 & 92.6 & 94.6 & 93.6 &    94.3 \\
Dog   &  93.3 & 95.1 & 92.7 & 94.4 & 95.2 & - & 90.0 & 92.6 & 93.8 & 94.5 &    93.5 \\
Frog  &  98.3 & 98.1 & 97.6 & 98.2 & 98.0 & 97.7 & - & 98.1 & 96.8 & 97.8 &    97.8 \\
Horse &  97.0 & 97.5 & 98.0 & 98.0 & 97.3 & 95.8 & 93.7 & - & 97.1 & 96.6 &    96.8 \\
Ship  &  97.0 & 96.2 & 96.4 & 97.4 & 97.2 & 97.1 & 96.0 & 97.5 & - & 97.5 &    96.9 \\
Truck &  97.2 & 95.6 & 97.0 & 97.1 & 96.3 & 97.1 & 95.3 & 95.9 & 96.2 & - &    96.4 \\
\hline
\end{tabular}
\caption{Confusion matrix of \model{}+'s AUROC on one-class classification over CIFAR-10. Each row and column represents normal and anomaly class respectively. ($\gamma_l=0.01, \gamma_p=0.00$)}
\end{minipage}
\end{table}

%% file: tables/full_res_p00_l005.tex
\begin{table}[H]
\centering
\begin{minipage}{0.90\textwidth}
\centering
\begin{tabular}{ccccccccccc|c}
\hline
\multicolumn{1}{r}{} & \multicolumn{1}{r}{Plane} & Car & Bird & Cat & Deer & Dog & Frog & Horse & Ship & Truck & Mean \\ \hline
Plane &   -  & 95.0 & 89.9 & 94.8 & 94.2 & 94.6 & 93.4 & 95.0 & 85.5 & 94.4 &    93.0 \\
Car   &    98.8 & - & 98.6 & 98.7 & 99.0 & 98.8 & 97.9 & 98.8 & 98.6 & 99.2 &    98.7 \\
Bird  &    89.0 & 91.8 & - & 93.7 & 93.1 & 92.1 & 89.2 & 94.1 & 90.9 & 94.4 &    92.0 \\
Cat   &    91.0 & 87.8 & 89.8 & - & 90.6 & 81.6 & 82.7 & 90.3 & 89.2 & 88.8 &    88.0 \\
Deer  &    95.5 & 95.7 & 96.1 & 95.9 & - & 95.1 & 93.5 & 95.8 & 94.7 & 95.7 &    95.3 \\
Dog   &    93.4 & 94.8 & 96.1 & 93.8 & 96.1 & - & 93.7 & 95.5 & 94.3 & 95.3 &    94.8 \\
Frog  &    97.8 & 97.9 & 98.3 & 98.4 & 98.5 & 98.6 & - & 98.6 & 98.0 & 98.4 &    98.3 \\
Horse &    97.9 & 97.3 & 98.5 & 98.2 & 98.5 & 97.4 & 97.6 & - & 97.6 & 97.1 &    97.8 \\
Ship  &    97.8 & 97.8 & 97.8 & 97.4 & 98.0 & 96.1 & 97.9 & 98.0 & - & 97.4 &    97.6 \\
Truck &    97.3 & 97.0 & 97.1 & 96.6 & 96.1 & 97.0 & 96.0 & 96.0 & 95.9 & - &    96.6 \\ 
\hline
\end{tabular}
\caption{Confusion matrix of \model{}+'s AUROC on one-class classification over CIFAR-10. Each row and column represents normal and anomaly class respectively. ($\gamma_l=0.05, \gamma_p=0.00$)}
\end{minipage}
\end{table}

%% file: tables/full_res_p00_l010.tex
\begin{table}[H]
\centering
\begin{minipage}{0.90\textwidth}
\centering
\begin{tabular}{ccccccccccc|c}
\hline
\multicolumn{1}{r}{} & \multicolumn{1}{r}{Plane} & Car & Bird & Cat & Deer & Dog & Frog & Horse & Ship & Truck & Mean \\ \hline
Plane &    - & 95.4 & 92.2 & 96.2 & 94.1 & 92.3 & 94.0 & 95.8 & 92.1 & 94.8 &    94.1 \\
Car   &    98.8 & - & 98.8 & 98.6 & 99.1 & 98.7 & 98.2 & 98.6 & 98.8 & 99.2 &    98.7 \\
Bird  &    90.5 & 93.6 & - & 93.8 & 94.3 & 92.2 & 90.5 & 94.2 & 93.6 & 93.6 &    92.9 \\
Cat   &    90.5 & 89.7 & 89.2 & - & 91.4 & 84.2 & 84.3 & 90.7 & 90.4 & 88.5 &    88.8 \\
Deer  &    95.7 & 95.7 & 96.3 & 95.6 & - & 94.9 & 94.2 & 96.7 & 94.7 & 95.7 &    95.5 \\
Dog   &    94.5 & 94.1 & 95.9 & 94.1 & 96.0 & - & 93.4 & 96.2 & 94.4 & 94.6 &    94.8 \\
Frog  &    98.1 & 97.7 & 98.5 & 98.7 & 98.5 & 98.4 & - & 98.7 & 98.1 & 97.8 &    98.3 \\
Horse &    97.4 & 97.8 & 98.6 & 98.4 & 98.5 & 97.2 & 97.7 & - & 97.3 & 97.0 &    97.8 \\
Ship  &    98.5 & 97.8 & 97.7 & 97.6 & 98.3 & 97.2 & 96.4 & 98.2 & - & 97.6 &    97.7 \\
Truck &    97.7 & 97.0 & 97.0 & 96.6 & 96.7 & 96.5 & 96.5 & 96.3 & 96.3 & - &    96.7 \\ \hline
\end{tabular}
\caption{Confusion matrix of \model{}+'s AUROC on one-class classification over CIFAR-10. Each row and column represents normal and anomaly class respectively. ($\gamma_l=0.1, \gamma_p=0.00$)}
\end{minipage}
\end{table}

%% file: tables/full_res_p005_l001.tex
\begin{table}[H]
\centering
\begin{minipage}{0.90\textwidth}
\centering
\begin{tabular}{ccccccccccc|c}
\hline
\multicolumn{1}{r}{} & \multicolumn{1}{r}{Plane} & Car & Bird & Cat & Deer & Dog & Frog & Horse & Ship & Truck & Mean \\ \hline
Plane &    - & 92.1 & 88.4 & 93.8 & 89.2 & 91.1 & 88.9 & 92.3 & 80.6 & 88.0 &    89.4  \\
Car   &    97.5 & - & 97.9 & 98.2 & 98.7 & 97.0 & 97.7 & 98.1 & 95.0 & 95.8 &    97.3  \\
Bird  &    87.2 & 92.3 & - & 83.3 & 84.5 & 86.1 & 79.7 & 92.0 & 88.4 & 92.3 &    87.3  \\
Cat   &    86.8 & 88.0 & 80.7 & - & 82.8 & 69.9 & 68.6 & 85.9 & 86.6 & 86.0 &    81.7  \\
Deer  &    93.8 & 94.7 & 92.2 & 92.5 & - & 91.1 & 87.6 & 91.9 & 93.3 & 91.9 &    92.1  \\
Dog   &    92.2 & 94.9 & 89.6 & 86.2 & 92.9 & - & 89.7 & 92.3 & 93.9 & 92.6 &    91.6  \\
Frog  &    96.5 & 97.2 & 95.7 & 93.8 & 95.9 & 95.5 & - & 98.2 & 96.4 & 97.6 &    96.3  \\
Horse &    96.1 & 95.5 & 95.8 & 94.1 & 93.3 & 93.5 & 93.4 & - & 94.3 & 96.0 &    94.7  \\
Ship  &    93.6 & 94.8 & 96.4 & 96.7 & 97.0 & 96.5 & 96.7 & 97.1 & - & 91.0 &    95.5  \\
Truck &    92.9 & 87.4 & 96.6 & 95.7 & 95.8 & 95.4 & 95.8 & 96.1 & 94.5 & - &    94.5  \\ \hline
\end{tabular}
\caption{Confusion matrix of \model{}+'s AUROC on one-class classification over CIFAR-10. Each row and column represents normal and anomaly class respectively. ($\gamma_l=0.01, \gamma_p=0.05$)}
\end{minipage}
\end{table}

%% file: tables/full_res_p005_l005.tex
\begin{table}[H]
\centering
\begin{minipage}{0.90\textwidth}
\centering
\begin{tabular}{ccccccccccc|c}
\hline
\multicolumn{1}{r}{} & \multicolumn{1}{r}{Plane} & Car & Bird & Cat & Deer & Dog & Frog & Horse & Ship & Truck & Mean \\ \hline
Plane &    - & 91.3 & 90.4 & 93.5 & 92.6 & 93.7 & 90.6 & 94.0 & 84.6 & 92.0 &    91.4 \\
Car   &    97.6 & - & 98.6 & 98.1 & 98.9 & 98.3 & 98.0 & 98.7 & 97.3 & 96.7 &    98.0 \\
Bird  &    88.6 & 90.3 & - & 86.5 & 89.3 & 83.2 & 82.2 & 90.7 & 90.7 & 92.4 &    88.2 \\
Cat   &    88.7 & 87.0 & 86.1 & - & 88.0 & 66.4 & 79.6 & 85.6 & 87.0 & 87.2 &    84.0 \\
Deer  &    94.4 & 94.8 & 93.7 & 92.1 & - & 93.8 & 90.8 & 91.0 & 93.6 & 94.1 &    93.1 \\
Dog   &    94.1 & 94.8 & 90.9 & 79.5 & 93.1 & - & 89.9 & 93.5 & 93.6 & 93.8 &    91.5 \\
Frog  &    97.5 & 96.6 & 93.7 & 95.6 & 97.8 & 95.3 & - & 98.1 & 97.6 & 97.5 &    96.7 \\
Horse &    96.7 & 96.3 & 96.7 & 94.0 & 96.1 & 92.4 & 96.7 & - & 96.3 & 96.5 &    95.8 \\
Ship  &    91.1 & 92.7 & 96.9 & 97.3 & 97.6 & 96.7 & 95.0 & 97.8 & - & 92.6 &    95.3 \\
Truck &    96.0 & 94.9 & 97.0 & 96.3 & 96.3 & 96.7 & 96.1 & 96.1 & 94.1 & - &    96.0 \\ \hline
\end{tabular}
\caption{Confusion matrix of \model{}+'s AUROC on one-class classification over CIFAR-10. Each row and column represents normal and anomaly class respectively. ($\gamma_l=0.05, \gamma_p=0.05$)}
\end{minipage}
\end{table}

%% file: tables/full_res_p005_l010.tex
\begin{table}[H]
\centering
\begin{minipage}{0.90\textwidth}
\centering
\begin{tabular}{ccccccccccc|c}
\hline
\multicolumn{1}{r}{} & \multicolumn{1}{r}{Plane} & Car & Bird & Cat & Deer & Dog & Frog & Horse & Ship & Truck & Mean \\ \hline
Plane &    - & 91.6 & 88.6 & 94.3 & 90.7 & 93.6 & 92.7 & 94.5 & 86.0 & 93.1 &    91.7 \\
Car   &    98.4 & - & 98.8 & 97.8 & 98.6 & 98.1 & 97.1 & 98.3 & 97.8 & 97.0 &    98.0 \\
Bird  &    90.4 & 91.2 & - & 89.7 & 86.3 & 85.3 & 84.8 & 91.6 & 89.5 & 92.7 &    89.0 \\
Cat   &    89.3 & 86.0 & 85.0 & - & 86.5 & 70.9 & 78.3 & 86.9 & 88.3 & 87.0 &    84.2 \\
Deer  &    94.9 & 94.6 & 89.4 & 90.4 & - & 92.8 & 90.7 & 93.8 & 95.1 & 94.4 &    92.9 \\
Dog   &    93.2 & 93.5 & 91.7 & 82.5 & 92.6 & - & 88.6 & 91.6 & 93.8 & 93.5 &    91.2 \\
Frog  &    97.4 & 97.0 & 95.5 & 92.7 & 96.4 & 95.6 & - & 97.7 & 97.5 & 97.7 &    96.4 \\
Horse &    96.4 & 96.5 & 95.8 & 95.3 & 95.8 & 94.1 & 97.2 & - & 96.0 & 96.0 &    95.9 \\
Ship  &    94.9 & 96.2 & 96.3 & 97.3 & 97.5 & 96.8 & 95.7 & 97.8 & - & 93.2 &    96.2 \\
Truck &    97.1 & 88.5 & 97.2 & 96.3 & 97.1 & 95.5 & 96.2 & 96.2 & 93.3 & - &    95.3 \\ \hline
\end{tabular}
\caption{Confusion matrix of \model{}+'s AUROC on one-class classification over CIFAR-10. Each row and column represents normal and anomaly class respectively. ($\gamma_l=0.1, \gamma_p=0.05$)}
\end{minipage}
\end{table}

%% file: tables/full_res_p010_l001.tex
\begin{table}[H]
\centering
\begin{minipage}{0.90\textwidth}
\centering
\begin{tabular}{ccccccccccc|c}
\hline
\multicolumn{1}{r}{} & \multicolumn{1}{r}{Plane} & Car & Bird & Cat & Deer & Dog & Frog & Horse & Ship & Truck & Mean \\ \hline
Plane &    - & 87.7 & 88.8 & 90.9 & 87.1 & 86.1 & 85.9 & 89.4 & 78.7 & 81.4 &    86.2 \\
Car   &    97.1 & - & 98.8 & 98.2 & 99.0 & 98.0 & 98.0 & 98.0 & 95.1 & 90.5 &    97.0 \\
Bird  &    87.8 & 91.1 & - & 86.2 & 81.8 & 80.2 & 76.8 & 85.6 & 90.2 & 91.7 &    85.7 \\
Cat   &    83.1 & 79.9 & 77.3 & - & 75.3 & 64.7 & 68.0 & 81.4 & 85.3 & 85.9 &    77.9 \\
Deer  &    93.6 & 93.2 & 91.1 & 84.7 & - & 88.7 & 86.4 & 86.9 & 93.6 & 92.7 &    90.1 \\
Dog   &    89.0 & 92.6 & 88.1 & 82.2 & 88.8 & - & 87.6 & 92.7 & 90.8 & 90.8 &    89.2 \\
Frog  &    96.4 & 96.2 & 89.0 & 88.3 & 89.5 & 93.4 & - & 97.9 & 95.2 & 97.3 &    93.7 \\
Horse &    96.2 & 96.3 & 94.7 & 93.3 & 85.5 & 91.9 & 94.6 & - & 96.0 & 93.8 &    93.6 \\
Ship  &    86.9 & 93.5 & 95.4 & 95.3 & 95.0 & 95.5 & 96.2 & 96.1 & - & 90.2 &    93.8 \\
Truck &    93.7 & 84.5 & 96.4 & 95.9 & 95.9 & 95.9 & 96.3 & 95.3 & 93.2 & - &    94.1 \\ \hline
\end{tabular}
\caption{Confusion matrix of \model{}+'s AUROC on one-class classification over CIFAR-10. Each row and column represents normal and anomaly class respectively. ($\gamma_l=0.01, \gamma_p=0.1$)}
\end{minipage}
\end{table}

%% file: tables/full_res_p010_l005.tex
\begin{table}[H]
\centering
\begin{minipage}{0.90\textwidth}
\centering
\begin{tabular}{ccccccccccc|c}
\hline
\multicolumn{1}{r}{} & \multicolumn{1}{r}{Plane} & Car & Bird & Cat & Deer & Dog & Frog & Horse & Ship & Truck & Mean \\ \hline
Plane &    - & 88.3 & 85.6 & 92.5 & 89.8 & 91.3 & 91.4 & 92.6 & 83.9 & 92.2 &    89.7 \\
Car   &    97.5 & - & 98.6 & 97.6 & 98.7 & 98.3 & 98.0 & 97.2 & 96.4 & 94.4 &    97.4 \\
Bird  &    83.5 & 90.6 & - & 82.6 & 83.3 & 79.7 & 82.1 & 87.2 & 87.4 & 92.2 &    85.4 \\
Cat   &    86.8 & 85.6 & 83.5 & - & 80.9 & 60.9 & 69.7 & 86.7 & 84.8 & 85.5 &    80.5 \\
Deer  &    93.3 & 94.9 & 89.7 & 88.4 & - & 88.3 & 86.0 & 88.4 & 93.4 & 93.3 &    90.6 \\
Dog   &    93.2 & 93.5 & 92.4 & 80.1 & 91.0 & - & 88.0 & 91.4 & 92.3 & 92.2 &    90.5 \\
Frog  &    96.5 & 96.2 & 93.9 & 90.7 & 92.1 & 90.6 & - & 97.5 & 96.5 & 97.4 &    94.6 \\
Horse &    94.5 & 96.3 & 95.9 & 92.1 & 91.0 & 89.2 & 94.5 & - & 96.2 & 95.8 &    93.9 \\
Ship  &    93.2 & 89.6 & 96.4 & 96.3 & 96.9 & 96.5 & 95.3 & 96.5 & - & 89.9 &    94.5 \\
Truck &    91.7 & 87.9 & 95.9 & 96.6 & 95.3 & 96.2 & 96.2 & 95.8 & 89.4 & - &    93.9 \\ \hline
\end{tabular}
\caption{Confusion matrix of \model{}+'s AUROC on one-class classification over CIFAR-10. Each row and column represents normal and anomaly class respectively. ($\gamma_l=0.05, \gamma_p=0.1$)}
\end{minipage}
\end{table}

%% file: tables/full_res_p010_l010.tex
\begin{table}[H]
\centering
\begin{minipage}{0.90\textwidth}
\centering
\begin{tabular}{ccccccccccc|c}
\hline
\multicolumn{1}{r}{} & \multicolumn{1}{r}{Plane} & Car & Bird & Cat & Deer & Dog & Frog & Horse & Ship & Truck & Mean \\ \hline
Plane &    - & 90.2 & 84.4 & 92.1 & 91.8 & 92.4 & 89.0 & 92.3 & 80.7 & 86.2 &    88.8 \\
Car   &    97.8 & - & 98.5 & 97.8 & 98.4 & 98.4 & 97.4 & 98.2 & 95.4 & 88.1 &    96.7 \\
Bird  &    89.3 & 91.6 & - & 86.9 & 80.3 & 83.9 & 79.0 & 88.2 & 89.5 & 91.4 &    86.7 \\
Cat   &    86.8 & 86.4 & 81.0 & - & 81.6 & 61.1 & 75.3 & 81.3 & 87.1 & 86.7 &    80.8 \\
Deer  &    94.4 & 94.6 & 92.3 & 87.9 & - & 85.8 & 85.9 & 91.8 & 93.7 & 93.7 &    91.1 \\
Dog   &    92.0 & 93.1 & 91.0 & 79.2 & 89.9 & - & 88.6 & 90.9 & 91.4 & 91.4 &    89.7 \\
Frog  &    97.2 & 95.4 & 93.5 & 93.8 & 90.4 & 93.3 & - & 96.4 & 96.1 & 97.4 &    94.8 \\
Horse &    95.3 & 96.2 & 96.3 & 95.1 & 93.5 & 91.1 & 95.2 & - & 96.4 & 94.9 &    94.9 \\
Ship  &    93.0 & 92.5 & 95.6 & 96.5 & 96.6 & 96.4 & 95.3 & 97.2 & - & 89.5 &    94.7 \\
Truck &    92.9 & 86.7 & 97.0 & 94.7 & 95.9 & 95.2 & 96.4 & 94.9 & 92.5 & - &    94.0 \\ \hline
\end{tabular}
\caption{Confusion matrix of \model{}+'s AUROC on one-class classification over CIFAR-10. Each row and column represents normal and anomaly class respectively. ($\gamma_l=0.1, \gamma_p=0.1$)}
\end{minipage}
\end{table}

%% file: tables/cont_res0.0.tex
\begin{table}[H]
\centering
\begin{minipage}{0.90\textwidth}
\centering
\begin{tabular}{ccccccccccc|c}
\hline
\multicolumn{1}{r}{} & \multicolumn{1}{r}{0} & 1 & 2 & 3 & 4 & 5 & 6 & 7 & 8 & 9 & Mean \\ \hline
Ours & 0.95 & 0.99 & 0.99 & 0.90 &0.86 & 0.89 & 0.84 & 0.95 & 0.90 & 0.97 &  0.93\\
CSI  & 0.98 & 0.98 & 0.99 & 0.97 & 0.93 & 0.99 & 0.88 & 0.99 & 0.96 & 0.88 & 0.95\\
\hline
\end{tabular}
\caption{ One-class classification over ImageNet-10 in the semi-supervised scenario. ($\gamma_l=0.1, \gamma_p=0.0$)}
\label{tab:imagenet_nocontam}
\end{minipage}
\end{table}

%% file: tables/cont_res_0.1.tex
\begin{table}[H]
\centering
\begin{minipage}{0.90\textwidth}
\centering
\begin{tabular}{ccccccccccc|c}
\hline
\multicolumn{1}{r}{} & \multicolumn{1}{r}{0} & 1 & 2 & 3 & 4 & 5 & 6 & 7 & 8 & 9 & Mean \\ \hline
Ours & 0.93 & 0.97 & 0.99 & 0.89 & 0.78 & 0.89 & 0.80 & 0.93 & 0.85 & 0.95 & 0.90\\
CSI  & 0.88 & 0.86 & 0.83 & 0.93 & 0.86 & 0.97 & 0.79 & 0.94 & 0.92 & 0.62 & 0.86\\
\hline
\end{tabular}
\caption{One-class classification over ImageNet-10 in the contamination scenario. ($\gamma_l=0.1, \gamma_p=0.1$)}
\label{tab:imagenet_contam}
\end{minipage}
\end{table}